\documentclass[lettersize,journal]{IEEEtran}
\usepackage{amsmath,amssymb,mathrsfs,amsfonts}
\DeclareMathOperator*{\argmin}{arg\,min} 
\usepackage{algorithmic}
\usepackage{algorithm}
\usepackage{array}
\usepackage{textcomp}
\usepackage{stfloats}
\usepackage{url}
\usepackage{verbatim}
\usepackage{graphicx}
\usepackage{cite}

\usepackage{caption}
\usepackage{subcaption}
\usepackage{booktabs}
\usepackage{multirow}
\usepackage{threeparttable}
\usepackage{indentfirst}
\usepackage{hyperref}
\usepackage{color}

\begin{document}

\title{A Spatio-Temporal Online Robust Tensor Recovery Approach for Streaming Traffic Data Imputation}



\author{
Yiyang Yang, Xiejian Chi, Shanxing Gao, Kaidong Wang\textsuperscript{*}, and Yao Wang\textsuperscript{*}
\thanks{Yiyang Yang, Xiejian Chi, Kaidong Wang and Yao Wang are with the Center for Intelligent Decision-making and Machine Learning, School of Management, Xi'an Jiaotong University, Xi'an, China. Shanxing Gao is with the Department of Marketing, School of Management, Xi'an Jiaotong University, Xi'an, China. } 
\thanks{\textsuperscript{*}Corresponding author: Kaidong Wang (E-mail: wangkd13@gmail.com), Yao Wang (E-mail: yao.s.wang@gmail.com).}
}

\maketitle

\begin{abstract}
Data quality is critical to Intelligent Transportation Systems (ITS), as complete and accurate traffic data underpin reliable decision-making in traffic control and management. Recent advances in low-rank tensor recovery algorithms have shown strong potential in capturing the inherent structure of high-dimensional traffic data and restoring degraded observations. However, traditional batch-based methods demand substantial computational and storage resources, which limits their scalability in the face of continuously expanding traffic data volumes. Moreover, recent online tensor recovery methods often suffer from severe performance degradation in complex real-world scenarios due to their insufficient exploitation of the intrinsic structural properties of traffic data.
To address these challenges, we reformulate the traffic data recovery problem within a streaming framework, and propose a novel online robust tensor recovery algorithm that simultaneously leverages both the global spatio-temporal correlations and local consistency of traffic data, achieving high recovery accuracy and significantly improved computational efficiency in large-scale scenarios. Our method is capable of simultaneously handling missing and anomalous values in traffic data, and demonstrates strong adaptability across diverse missing patterns. Experimental results on three real-world traffic datasets demonstrate that the proposed approach achieves high recovery accuracy while significantly improving computational efficiency by up to three orders of magnitude compared to state-of-the-art batch-based methods. These findings highlight the potential of the proposed approach as a scalable and effective solution for traffic data quality enhancement in ITS.
\end{abstract}


\section{Introduction}
\subsection{Motivations}
\IEEEPARstart{W}{ith} the accelerating integration of digital and intelligent technologies into daily life, Intelligent Traffic Systems (ITS) are playing an increasingly vital role in modern urban management \cite{zhu2018big}. 
By collecting real-time measurements such as traffic flow and vehicle speed from cameras, sensors, and loop detectors, ITS enable energy-efficient traffic control, alleviate congestion, and facilitate infrastructure planning \cite{chen2017achieving,jin2023spatio,asl2024cycles}. 
Accordingly, the ability to obtain complete, accurate, and timely traffic data is fundamental to various downstream tasks, including traffic signal optimization \cite{zhang2021optimization}, route planning \cite{xiao2024smart}, congestion mitigation \cite{cheng2020mitigating}, and traffic prediction \cite{yu2025clear, miao2025spatio, zheng2019real}.

However, several practical challenges hinder the effective utilization of traffic data in ITS.
One of the most pervasive issues is missing data, which commonly results from sensor malfunctions, communication failures, or hardware degradation \cite{chen2019missing, hou2023missii, zhang2024comprehensive, zhang2025missing}.
As ITS analytics typically rely on complete data, missing values can significantly compromise the reliability of traffic monitoring and decision-making.
In addition, traffic data streams are frequently contaminated by outliers caused by sensor aging, extreme weather, or traffic incidents. 
If not properly addressed, such anomalies can severely distort estimations and degrade system performance, highlighting the need for robust outlier detection and elimination methods \cite{guo2015real,djenouri2019survey,wang2020anomaly}.

Early studies primarily addressed these issues using interpolation-based methods, owing to their simplicity and low computational cost. However, such approaches typically overlook the intrinsic spatio-temporal dependencies in traffic data, making them inadequate for reconstructing complex global patterns.
To overcome this limitation, matrix-based methods were introduced, where traffic data are represented as two-dimensional matrices (e.g., \emph{location} $\times$ \emph{time}) and recovered via matrix completion techniques that exploit their approximate low-rank structure. By leveraging shared patterns across sensors and over time, these approaches significantly improve recovery accuracy and highlight the benefits of incorporating structural priors over treating each sensor independently.

Nevertheless, real-world traffic systems often exhibit richer spatial and temporal regularities that cannot be fully captured by a simple matrix representation, leading to suboptimal recovery performance \cite{asif2016matrix}.
This motivates the use of tensor-based representations as a more expressive alternative for modeling the complex spatio-temporal structure inherent in traffic data.
For example, vehicular speed data collected over a week can be naturally organized as a third-order tensor with dimensions \emph{timestamp} $\times$ \emph{location} $\times$ \emph{day}.
On this basis, robust tensor recovery methods have been developed to simultaneously handle missing data and outliers by exploiting the intrinsic low-rank structure of tensors \cite{liu2012tensor, lu2018exact}. Compared to matrix-based approaches, tensor completion is more effective at capturing high-order structural correlations in traffic data (e.g., daily periodicity shared across locations), thereby typically achieving better recovery performance.

Although low-rank tensor completion methods have achieved notable success in traffic data recovery, they still face limitations in terms of computational efficiency and recovery accuracy when applied to complex, real-world ITS scenarios. These limitations are primarily reflected in the following two aspects.

On the one hand, in real-world ITS, traffic data are continuously generated from large-scale sensor networks in a streaming fashion and timely decision-making is essential \cite{chan2023missing, zhang2024comprehensive}. Most existing methods operate in an offline (batch) manner, typically processing the entire historical dataset at once and requiring full recomputation whenever new data arrive. This results in substantial computational and memory overhead, significantly limiting their computational efficiency and scalability in real-world ITS scenarios. With the continuous growth of traffic volumes and the increasing real-time demands of modern ITS applications, it becomes critical to develop online recovery methods that can incrementally process new data without accessing or recomputing the entire dataset.

On the other hand, in real-world ITS, traffic data often exhibit highly complex and diverse missing patterns.  These can range from high missing rates to structured data loss, such as long consecutive time intervals with no readings or entire spatial regions of the network going offline due to sensor failure or communication issues. Such challenging scenarios significantly increase the difficulty of accurate data recovery. Conventional tensor recovery approaches primarily rely on global low-rank assumptions, which are often insufficient to handle these complex missing scenarios. This has motivated growing interest in incorporating additional structural priors to improve recovery robustness \cite{chen2019bayesian,chen2020nonconvex,lyu2024tucker}. 
In practice, traffic data exhibit strong \textit{local consistency} in both spatial and temporal dimensions. For example, neighboring intersections tend to experience similar traffic conditions, and traffic states in adjacent time periods are usually continuous and smooth. Effectively capturing this fine-grained local consistency can provide strong guidance for recovering missing or corrupted values, especially under severe or structured missingness, thereby enhancing the reliability of ITS in complex, real-world environments. Therefore, effectively incorporating both spatio-temporal global correlations and local consistency into tensor completion modeling, particularly in an online setting, is crucial for enhancing the accuracy of traffic data recovery. 

To overcome these limitations, we propose a novel online robust tensor recovery approach for streaming traffic data imputation, which incrementally processes spatio-temporal data as it arrives. To preserve global spatio-temporal dependencies in an online setting, we factorize the streaming traffic tensor (e.g., \emph{timestamp} $\times$ \emph{location} $\times$ \emph{day}) into temporally evolving factor matrices and a shared core tensor, maintaining a compact low-rank structure in real time. To further improve recovery performance, we incorporate spatial and temporal regularizations along the location and timestamp dimensions, respectively. These constraints guide the model to capture fine-grained spatio-temporal local consistency, leading to more accurate imputation and more robust anomaly detection. 
Ultimately, we develop an online tensor recovery model that integrates global spatio-temporal correlations with local consistency priors, and devise an efficient algorithm to solve it in a streaming manner. Experimental results on three real-world traffic datasets demonstrate that the proposed method consistently achieves superior recovery accuracy and computational efficiency compared to state-of-the-art baselines, particularly in challenging scenarios involving low sampling rates and structured missing patterns.
This significant computational advantage makes it well-suited for modern ITS applications that require timely responses to continuously streaming traffic data.

\subsection{Related work}
Early approaches to recovering corrupted spatio-temporal traffic data primarily relied on regression-based analysis, which interpolated missing values by modeling their relationships with observed data in the spatial and temporal domains. Typical methods include multivariate regression models \cite{raghunathan2001multivariate, chen2003detecting}, the Auto-Regressive Integrated Moving Average (ARIMA) model \cite{sharma2004effect, zhong2004genetically}, among others. While these methods are effective for capturing simple temporal trends or localized spatial dependencies, they often fall short in modeling the complex, high-dimensional, and nonlinear spatio-temporal patterns present in real-world traffic data. To better exploit the underlying structure of traffic data, low-rank matrix and tensor completion methods have become mainstream in subsequent research, which assume that the observed data lie near a low-dimensional subspace and can thus be effectively reconstructed from partial observations \cite{wu2020imputation}. In terms of processing strategy, existing matrix and tensor recovery methods are generally divided into batch-based and online methods. We briefly review both lines of work below.

\subsubsection{Batch-based Methods}
Most existing low-rank matrix and tensor completion methods operate in a batch setting, where the entire dataset is assumed to be available beforehand. A considerable body of empirical research suggests that matrix decomposition methods are more effective than regression-based techniques for recovering incomplete traffic data \cite{asif2016matrix, chen2017ensemble}. Among them, Luo et al. \cite{luo2019traffic} proposed the ILRMD model, which leverages the spatio-temporal characteristics of traffic flow data and achieves strong performance. Sure et al. \cite{sure2021spatio} further improved recovery accuracy by incorporating spatio-temporal correlation terms into traditional matrix decomposition models, thereby capturing the intrinsic structure of traffic flow more effectively. More recently, Chen et al. \cite{chen2024laplacian} introduced the Laplacian Convolutional Representation (LCR) model which integrates circulant matrix nuclear norm with Laplacian kernel-based spatio-temporal regularization, achieving accurate imputation under noisy and sparse conditions.

Despite their effectiveness, matrix-based methods are inherently limited in capturing higher-order dependencies in traffic data. To address this, tensor-based approaches have been proposed, which model traffic data as multi-dimensional arrays and better exploit complex spatio-temporal correlation structures for improved recovery performance. In the context of ITS, tensor recovery algorithms primarily differ in the decomposition strategies used to exploit the low-rank structure of traffic data.
For example, \cite{asif2013low} and \cite{tan2013tensor} applied  CANDECOMP/PARAFAC (CP) and Tucker decompositions, respectively, to model the underlying structure of traffic tensors for recovery tasks. Subsequently, probabilistic extensions such as Bayesian Augmented Tensor Factorization (BATF) \cite{chen2019missing} and Bayesian Gaussian CP (BGCP) \cite{chen2019bayesian} were proposed to enhance robustness in multivariate traffic time series imputation. \cite{chen2020nonconvex} performed traffic data imputation under the T-SVD decomposition framework by minimizing a non-convex truncated nuclear norm as the optimization objective. \cite{lyu2024tucker} integrated time-series decomposition with joint Tucker factorization and rank minimization, enabling robust recovery from complex missing patterns and outliers, without the need for exhaustive rank tuning.

\subsubsection{Online Methods}
Batch-based recovery methods must reprocess the entire historical data upon the arrival of new observations, resulting in significant computational overhead.  This limitation has motivated the development of online methods that can incrementally update the model as new traffic data streams in. For instance, \cite{he2011online} proposed GRASTA, an online matrix recovery method that leverages Grassmannian geometry and $l_1$-norm optimization to robustly track low-rank subspaces in the presence of sparse outliers. Similarly, \cite{shen2016online} developed OLRSC, an online extension of Low-Rank Representation (LRR) that reduces memory usage while ensuring efficient matrix recovery for large-scale datasets. More recently, \cite{dung2021robust} proposed an online method that integrates outlier rejection via ADMM with incremental subspace estimation (PETRELS-ADMM), effectively handling both missing data and outliers.

For online tensor recovery, \cite{mardani2015subspace} proposed TeCPSGD, which employs stochastic gradient descent to track the CP decomposition of third-order streaming tensors with missing entries. \cite{sobral2015online} proposed an online low-rank tensor recovery method based on a stochastic tensor decomposition framework (OSTD), while \cite{wu2022online} extended the matrix-based OLRSC algorithm to tensor structures within the T-SVD framework (OLRTSC). To improve recovery performance, \cite{zhou2019multi} developed a coupled tensor completion method, while \cite{nie2022truncated} defined a new tensor paradigm and applied a corresponding low-rank completion model for missing traffic flow data estimation. Subsequently, \cite{abed2022robust} proposed a robust adaptive CP decomposition (RACP) method to address the challenges of high-order incomplete streaming tensors contaminated by outliers. \cite{wu2025cauchy} proposed CL-NLFT, a non-negative CP-based tensor completion model that employs the Cauchy loss instead of the conventional L2 loss to enhance robustness against outliers in traffic data. In the related domain of network traffic, \cite{kasai2016network} developed an online CP-based subspace tracking method using a Hankelized time-structured traffic tensor to model normal flow patterns. \cite{kasai2019fast} proposed OLSTEC, a fixed-rank tensor completion algorithm based on CP decomposition and recursive least squares. \cite{zhang2018variational} proposed a Bayesian Robust Streaming Tensor Factorization (BSTF) model that automatically infers the underlying tensor rank to accurately capture low-rank structures, and applied it to the reconstruction and completion of network traffic data.

Our method differs from previous online tensor recovery approaches in two key aspects.\\
(1) From a modeling perspective, we are the first to explicitly incorporate both spatio-temporal global correlations and local consistency into an online robust recovery framework for traffic data. This comprehensive exploitation of prior structural information ensures that our model achieves high recovery accuracy while significantly improving computational efficiency.\\
(2) From an application perspective, our method demonstrates strong adaptability to various complex scenarios commonly encountered in real-world ITS, such as high missing rates, structured missing patterns, and the presence of outliers. This robustness significantly broadens the applicability of the algorithm, making it more suitable for practical deployment in large-scale, dynamic traffic environments.

\subsection{Challenges and contributions}
Despite the growing interest in robust tensor recovery for streaming traffic data, there remain several fundamental challenges that have yet to be effectively tackled.
First, existing methods often fall short in fully exploiting the inherent structural priors of traffic flow data. Most approaches focus predominantly on modeling global low-rank structures, while overlooking the local consistency that is intrinsic to urban traffic networks along both spatial and temporal dimensions \cite{dell2015time}. Second, existing approaches often struggle to maintain consistently reliable recovery performance when faced with the complex and diverse scenarios common in real-world ITS \cite{tan2013tensor, asif2016matrix}.  Scenarios such as high missing ratios, structured missing patterns, and outlier contamination greatly complicate the recovery process and often lead to significant performance degradation. This, in turn, limits the practical applicability and reliability of existing methods in real-world ITS environments, where robustness across diverse conditions is critical.

To address these challenges, we propose a novel online robust tensor recovery framework for streaming spatio-temporal traffic data imputation. By fully exploiting both the global and local intrinsic structures of traffic data in an online manner, our method significantly improves computational efficiency while maintaining high recovery accuracy. More importantly, it consistently delivers robust and reliable performance under complex and diverse conditions, thereby greatly enhancing its practical applicability in real-world ITS environments. Our work makes three primary contributions:
\begin{itemize}
    \item We propose the first online robust tensor recovery framework that explicitly integrates both global low-rank structure and local spatio-temporal consistency, enabling more accurate traffic data recovery in complex real-world scenarios. 
    \item We design a scalable and memory-efficient algorithm that incrementally updates the tensor decomposition as new data arrives, significantly reducing computational overhead compared to traditional batch-based methods.
    \item We conduct comprehensive experiments on three large-scale real-world traffic datasets under varying missing rates and diverse missing patterns. Experimental results demonstrate that the proposed approach achieves higher recovery accuracy than existing streaming imputation methods, particularly under complex real-world scenarios with high missing rates and structured missing patterns, while delivering up to three orders of magnitude improvement in computational efficiency compared to batch-based algorithms. 
\end{itemize}

\section{Preliminaries}

\subsection{Notations}
Throughout this work, we use the following notational conventions. Matrices are indicated with bold uppercase letters, e.g., $\mathbf{X} \in \mathbb{R}^{m \times n}$, with the $r$-th row of matrix $\mathbf{X}$ denoted by $\mathbf{X}[r,:]$, and the $(i,j)$-th entry of matrix $\mathbf{X}$ denoted by $\mathbf{X}[i,j]$. Vectors and scalars are denoted by bold and non-bold lowercase letters, respectively, e.g., $\mathbf{x} \in \mathbb{R}^n$ and $x$. We use $\mathbf{X}^{-1}$, $\mathbf{X}^{\dagger}$, $\mathbf{X}^{\top}$ and $tr(\mathbf{X})$ to represent the inverse, the pseudo-inverse, the transpose, and the trace of $\mathbf{X}$, respectively. The Frobenius norm of a matrix $\mathbf{X}$ is calculated as $\|\mathbf{X}\|_F=\sqrt{\sum_{i, j} {\mathbf{X}[i,j]}^2}$. We also explore third-order tensor, represented as $\mathcal{X} \in \mathbb{R}^{n_1 \times n_2 \times n_3}$, and define the Frobenius norm similarly: $\|\mathcal{X}\|_F=\sqrt{\sum_{i_1, i_2, i_3} {\mathcal{X}[i_1,i_2,i_3]}^2}$. Furthermore, the unfolding of tensor $\mathcal{X}$ along its $k$-th mode is denoted by $\mathcal{X}^{(k)} \in \mathbb{R}^{n_k \times\left(\Pi_{l \neq k} n_l\right)}$, and the $t$-th frontal slice of tensor $\mathcal{X}$ is expressed as $\overrightarrow{\mathcal{X}_{t}}\in \mathbb{R}^{n_1 \times n_2}$. The operations of element-wise multiplication, Kronecker product, and concatenation for tensors are denoted by $\circledast$, $\otimes$ and $\boxplus$, respectively.

\subsection{Problem Definition}\label{section: Problem Definition}
In this work, we introduce the problem of recovering streaming spatio-temporal traffic data in the presence of missing values and outliers in a general sense. Streaming spatio-temporal traffic data refers to the continuous acquisition of traffic data collected by various sensors over time. Such data is inherently dynamic, with its volume continuously increasing over time as new measurements are collected. To effectively model this complex and evolving dataset, we represent the traffic data collected over the first $t$ days as an incomplete streaming third-order tensor $\mathcal{M}_t \in \mathbb{R}^{n_1 \times n_2 \times t}$, where the dimensions correspond to $timestamp \times location \times day$, respectively. Each element within this tensor represents traffic data at a specific timestamp and location, collected over successive days. As new data $\overrightarrow{\mathcal{M}}_{t+1}\in \mathbb{R}^{n_1 \times n_2}$ is collected at $(t+1)$-th day, it is integrated into the dataset as a frontal slice of the tensor $\mathcal{M}_{t+1}$, i.e. $\mathcal{M}_{t+1} = \mathcal{M}_{t} \boxplus\overrightarrow{\mathcal{M}}_{t+1}$. This approach preserves the consistency of the first two dimensions (timestamp and location), while allowing the third dimension, representing days, to expand incrementally. 

\begin{figure}[!t]
  \centering
  \begin{subfigure}[b]{0.21\textwidth}
    \centering
    \includegraphics[width=\linewidth, height=0.15\textheight]{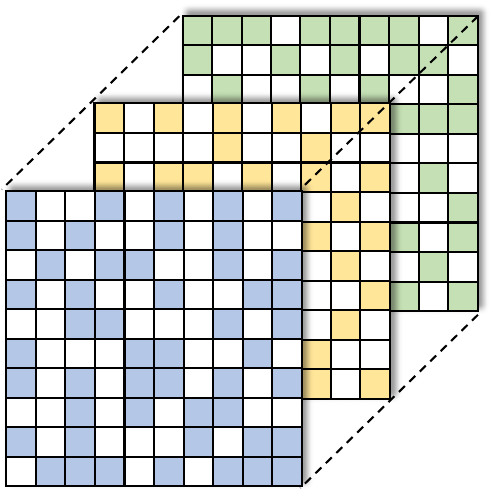}
    \caption{RM}\label{mp:sub1}
  \end{subfigure}\hspace{0.02\linewidth} 
  \begin{subfigure}[b]{0.21\textwidth}
    \centering
    \includegraphics[width=\linewidth, height=0.15\textheight]{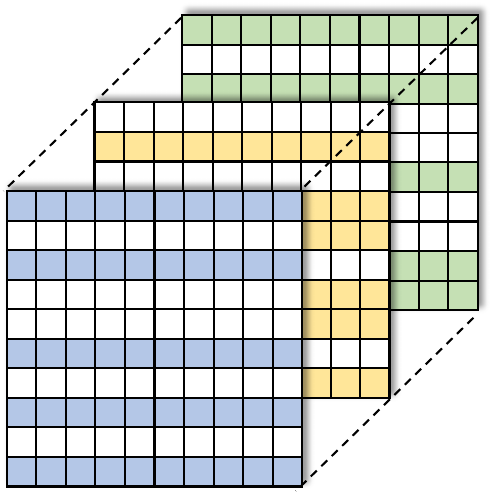}
    \caption{TM}\label{mp:sub2}
  \end{subfigure}
  
  \begin{subfigure}[b]{0.21\textwidth}
    \centering
    \includegraphics[width=\linewidth, height=0.15\textheight]{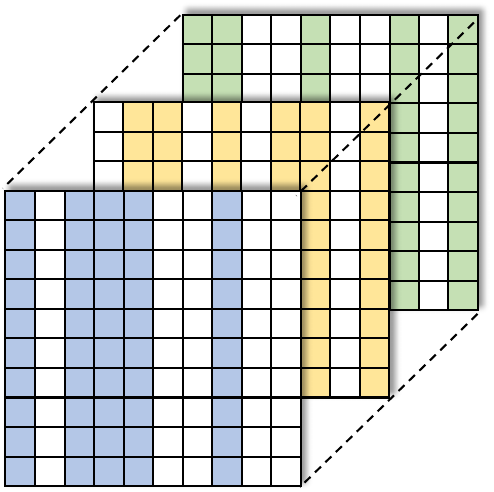}
    \caption{SM}\label{mp:sub3}
  \end{subfigure}
  \hspace{0.02\linewidth}
  \begin{subfigure}[b]{0.21\textwidth}
    \centering
    \includegraphics[width=\linewidth, height=0.15\textheight]{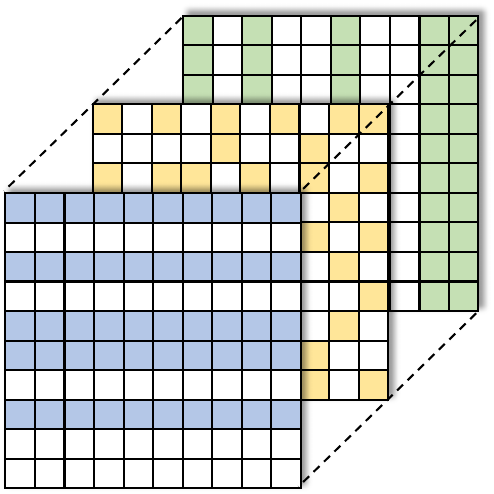}
    \caption{MM}\label{mp:sub4}
  \end{subfigure}
  \caption{Distinct categories of missing patterns}
  \label{missing_pattern}
\end{figure}

Considering the diversity of missing data scenarios in real-world traffic applications, we identify four representative categories of missing patterns. A binary mask tensor $ \mathcal{P}_t \in \mathbb{R}^{n_1 \times n_2 \times t} $ is introduced to represent the missing structure of data tensor $ \mathcal{M}_t $. Specifically, the $ t $-th frontal slice of $ \mathcal{P}_t $, $\overrightarrow{\mathcal{P}_t}$, represents the missing pattern on $ t $-th day, where
$\overrightarrow{\mathcal{P}_t}[i_1, i_2] = 1$ indicates that the entry $\overrightarrow{\mathcal{M}_t}[i_1, i_2]$ is observed, and $\overrightarrow{\mathcal{P}_t}[i_1, i_2] = 0$ otherwise. The four categories of missing patterns are described as follows:

\begin{itemize}
    \item Random Missing (RM): The daily traffic data are randomly missing, i.e. the elements of the mask $\overrightarrow{\mathcal{P}_t}$ are randomly set to 0 at each $t$-th day, as shown in Figure \ref{mp:sub1};
    \item Temporal Missing (TM): The daily traffic data are missing in the temporal dimension, i.e. the elements of the whole row of the mask $\overrightarrow{\mathcal{P}_t}$ are randomly set to 0 at each $t$-th day, as shown in Figure \ref{mp:sub2};
    \item Spatial Missing (SM): The daily traffic data are missing in the spatial dimension, i.e. the elements of the whole column of the mask $\overrightarrow{\mathcal{P}_t}$ are randomly set to 0 at each $t$-th day, as shown in Figure \ref{mp:sub3};
    \item Mix Missing (MM): The pattern of missing traffic data varies from day to day, i.e., the missing pattern of the mask $\overrightarrow{\mathcal{P}_t}$ at different $t$-th day is Temporal Missing, Spatial Missing, or Spatial Missing with equal probability, as shown in Figure \ref{mp:sub4}.
\end{itemize}

Considering the inherent low rank structure of traffic tensors and the presence of outliers, the data tensor $\mathcal{M}_t$ can be decomposed into the sum of two components:
\begin{equation}\label{equ_1}
\mathcal{M}_t = \mathcal{N}_t + \mathcal{S}_t,
\end{equation}
where $\mathcal{N}_t$ denotes the underlying low-rank tensor that captures the spatio-temporal global correlations, and $\mathcal{S}_t$ represents a sparse outlier tensor corresponding to anomalies or corruptions. To enhance data recovery from streaming spatio-temporal traffic tensor, we introduce a refined online tensor decomposition framework tailored to real-time and robust traffic data analysis. The proposed framework processes streaming data in an online manner, significantly reducing computational and memory overhead. Meanwhile, by incorporating carefully designed regularization terms, it effectively captures spatio-temporal fine-grained local consistency, leading to more accurate and robust data recovery.
In this framework, the $ t $-th frontal slice of the low-rank component $  \mathcal{N}_t $, denoted as $\overrightarrow{\mathcal{N}_t}$, is represented using Tucker decomposition with Tucker rank $(r_1, r_2, r_3)$ as follows:
\begin{equation}\label{equ_Tucker}
    \overrightarrow{\mathcal{N}_t} = \mathcal{G} \times_{1} \mathbf{U}_{T} \times_{2} \mathbf{U}_{S} \times_3 {\mathbf{u}_{D,t}}^{\top},
\end{equation}
where $\mathcal{G} \in \mathbb{R}^{r_1 \times r_2 \times r_3}$ denotes the core tensor, while $\mathbf{U}_T \in \mathbb{R}^{n_1 \times r_1}$ and $\mathbf{U}_S \in \mathbb{R}^{n_2 \times r_2}$ represent the temporal and spatial factor matrices, respectively. The vector $\mathbf{u}_{D,t} \in \mathbb{R}^{r_3}$ acts as a dynamic weight vector associated with day $t$. In dynamic environments, the core tensor and factor matrices, denoted as $\mathcal{G}_t$, $\mathbf{U}_{T,t}$, and $\mathbf{U}_{S,t}$, may vary slowly over time to capture gradual changes in data structure.

\begin{figure*}[htbp]
  \centering
  \includegraphics[width=0.7\linewidth]{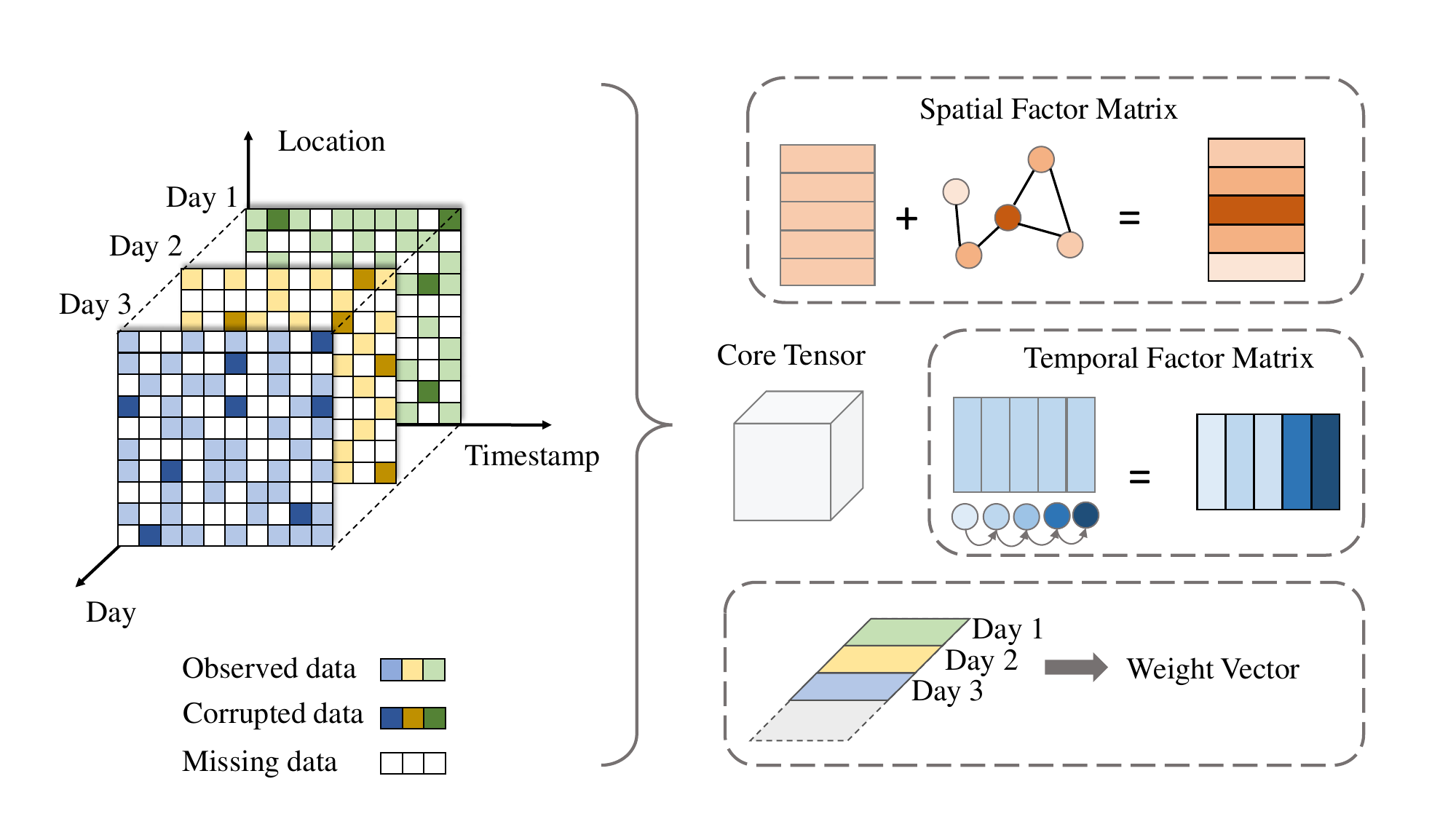}
  \caption{Illustration of the proposed online robust tensor decomposition framework}
  \label{fw}
\end{figure*}

\section{Model}
\subsection{Spatial Local Consistency}
To better capture the fine-grained structure of traffic data, we first model its spatial local consistency, which refers to the tendency of nearby locations to exhibit similar traffic behavior. This property is particularly evident in urban road networks, where adjacent intersections or connected segments often experience correlated traffic states \cite{wang2018traffic}. To quantify the proximity among locations, we first define a spatial graph where each node represents a sensor or location, and the edge weights encode spatial closeness or connectivity. This graph structure enables the model to explicitly capture spatial correlations based on physical distance or road network topology. 

To this end, we define a graph with $N$ nodes, where each node corresponds to a traffic sensor and is associated with a vector $\mathbf{y}_i$ representing all traffic data recorded on a specific day. The graph edges are weighted to reflect spatial closeness between sensors, i.e., larger weights denote smaller physical distances or stronger spatial connectivity. In practice, various strategies exist for defining edge weights in spatial graphs, including binary $(0–1)$ weighting, heat kernel weighting, and dot-product weighting \cite{cai2010graph}.  In this work, we adopt the Gaussian weighting scheme due to its effectiveness and widespread use in graph-based data analysis \cite{belkin2001laplacian, kalofolias2016learn}. The weight between nodes i and j is defined as:
$$
    \mathbf{W}[i,j]=e^{-\left(\left\|\mathbf{y}_i-\mathbf{y}_j\right\|^2\right) / \sigma^2},
$$
where $\mathbf{y}_i$ and $\mathbf{y}_j$ represent traffic time series recorded at nodes $i$ and $j$, and $\sigma$ is a bandwidth parameter controlling the decay rate of similarity with distance.

Based on the weight matrix $\mathbf{W}$, we further define the Laplace matrix $\mathbf{L}$ to encode the local similarity of the traffic data in the spatial dimension:
$$
    \mathbf{L}=\mathbf{D}-\mathbf{W},
$$
where $\mathbf{D} \in \mathbb{R}^{N \times N}$ is a diagonal degree matrix with elements $\mathbf{D}[i,i]=\sum_{j=1}^{N} \mathbf{W}[i,j]$, $i=1, \ldots, N$. Consequently, we formulate the spatial regularization term to capture the local consistency of traffic data in the spatial dimension as:
\begin{equation*}\label{equ_spa}
    \sum_{i=1}^{N} \sum_{j=1}^{N} \mathbf{W}[i,j]\left\|\mathbf{U}_{S}[i,:]-\mathbf{U}_{S}[j,:]\right\|_F^2=\operatorname{tr}\left({\mathbf{U}_{S}}^{\top} \mathbf{L} \mathbf{U}_{S}\right),
\end{equation*}
where $\mathbf{U}_S$ denotes the spatial factor matrix.

\subsection{Temporal Local Consistency}
We then model the temporal local consistency of traffic data, which reflects the observation that traffic conditions tend to evolve in a gradual and continuous manner over time. In real-world scenarios, the state of traffic at a given timestamp is usually similar to that at adjacent timestamps, unless disrupted by anomalies such as accidents or extreme weather events. By capturing this gradual temporal evolution, the model can better recover missing or corrupted values while maintaining the natural continuity of traffic dynamics. 
This temporal consistency can be effectively characterized by the Toeplitz operator \cite{zhou2016robust,deng2021graph}, defined as follows:
$$
\operatorname{Toeplitz(-1,1,0)}=\left(\begin{array}{cccccc}
-1 & 1 & 0  & \cdots & 0 \\
0 & -1 & 1  & \cdots & 0\\
\vdots & \vdots & \vdots & \ddots &\vdots \\
0 & 0 &  0 & \cdots & 1 \\
1 & 0 &  0 & \cdots & -1 
\end{array}\right)_{n \times n.}
$$

Specifically, for the temporal factor matrix $\mathbf{U}T$ of the traffic tensor, smoothness is encouraged by penalizing the differences between adjacent rows, leading to the following constraint:
\begin{equation}\label{equ_tem}
     \sum_{i=2}^{n}\left\|\mathbf{U}_{T}[i,:]-\mathbf{U}_{T}[i-1,:]\right\|_F^2=\left\|\mathcal{T} \mathbf{U}_{T}\right\|_F^2, 
\end{equation}
where $\mathcal{T} = \operatorname{Toeplitz(-1,1,0)}$ and $\mathbf{U}_{T}[i,:]$ denotes the $ i $-th row vector of $\mathbf{U}_{T} $. Consequently, we use $\|\mathcal{T} \mathbf{U}_{T}\|_F^2$ as the regularization term to characterize the temporal local consistency of the traffic tensor.

\subsection{Spatial-Temporal Online Robust Tucker Decomposition Model}
Based on the above regularization terms, we formulate an online robust tensor recovery model under the Tucker decomposition framework to efficiently process streaming traffic data in real time. The foundation of our model is based on the decomposition formulations presented in Equation (\ref{equ_1}) and (\ref{equ_Tucker}).
At each time step $t$, we receive a new observed tensor slice $\vec{\mathcal{M}}_{t}$ (with missing entries), and aim to adaptively update its dynamic factor $\mathbf{u}_{D,t}$ and the sparse outlier $\vec{\mathcal{S}}_{t}$ based on the current observation. Meanwhile, considering the potential gradual evolution of subspaces in both spatial and temporal dimensions under dynamic traffic environments, we simultaneously update the core tensor $\mathcal{G}_t$ and factor matrices $\mathbf{U}_{T,t}$, and $\mathbf{U}_{S,t}$ to better track such changes. Then the complete traffic data at time step $t$ can be reconstructed as the estimated low-rank tensor slice: $$\overrightarrow{\mathcal{N}_{t}} \leftarrow \mathcal{G}_{t} \times_{1} \mathbf{U}_{T,t} \times_{2} \mathbf{U}_{S,t} \times_3 {\mathbf{u}_{D,t}}^{\top}.$$
This frame-by-frame recovery mechanism allows the model to dynamically track the evolving low-rank structure in real time, without relying on batch optimization over the entire tensor. 

To achieve this, we minimize an exponentially weighted objective function $\mathbb{F}(\cdot)$ that incorporates spatial and temporal regularization terms, as defined below:
\begin{align}\label{equ_3}
    \mathbb{F}(\mathcal{G}_t,\mathbf{U}_{S,t},\mathbf{U}_{T,t}) 
    \triangleq &\min_{\mathcal{G},\mathbf{U}_S,\mathbf{U}_T} \ 
    \sum_{k=1}^t \lambda^{t-k} I(\mathcal{G},\mathbf{U}_S,\mathbf{U}_T;\vec{\mathcal{P}}_k, \vec{\mathcal{M}}_k) \nonumber \\
    &~~~~+ \alpha \operatorname{tr}(\mathbf{U}_S^{\top} \mathbf{L} \mathbf{U}_S) 
    + \beta \|\mathcal{T} \mathbf{U}_T\|_F^2
\end{align}
where the loss function $I(\cdot)$ with respect to the $k$-th slice $\vec{\mathcal{M}}_k$ is given by:
\begin{align}\label{equ_4}
&I\left(\mathcal{G},\mathbf{U}_{S},\mathbf{U}_{T};\vec{\mathcal{P}}_k, \vec{\mathcal{M}}_k\right)
    \stackrel{\Delta}{=} \min_{\mathbf{u}_{D},\vec{\mathcal{S}}} 
    \Big\| \vec{\mathcal{P}}_k \circledast \nonumber\\
    &\Big(\vec{\mathcal{M}}_k - \mathcal{G} \times_{1} \mathbf{U}_{T} \times_{2} \mathbf{U}_{S} \times_3 \mathbf{u}_{D,k}^{\top}  - \vec{\mathcal{S}}_{k}\Big)\Big\|_F^2 + \gamma \|\vec{\mathcal{S}}_k\|_1,
\end{align}
and $\lambda \in [0,1]$ is the forgetting factor that  applies exponential decay to past observations, ensuring recent data has a stronger influence on the recovery process.
$\mathbf{L}$ and $\mathcal{T}$ represent the graph Laplacian matrix and Toeplitz operator, respectively, to incorporate spatial and temporal local consistency constraints with regularization parameters $\alpha$ and $\beta$. $\left|\cdot\right|_1$ denotes the $\ell_1$-norm used to measure the magnitude of sparse outliers, and $\gamma$ is the regularization parameter that controls the strength of the sparsity constraint.

In conclusion, the model defined in Equation \ref{equ_3} is referred to as the Spatio-Temporal Online Robust Tucker Decomposition (STORTD), which enables robust and efficient processing of dynamic traffic data in streaming environments.

\section{Algorithm}

\subsection{Estimation of outlier $\overrightarrow{\mathcal{S}_{t}}$ and weight $\mathbf{u}_{D,t}$}

We can derive the outlier $\overrightarrow{\mathcal{S}_{t}}$ and the weight vector $\mathbf{u}_{D,t}$ from the minimization of $ I\left(\mathcal{G},\mathbf{U}_{S},\mathbf{U}_{T},  \overrightarrow{\mathcal{P}_k}, \overrightarrow{\mathcal{M}_k}\right)$ in Equation \ref{equ_4} as follows:
\begin{small}
\begin{equation}\label{equ_5}
    \{\vec{\mathcal{S}}_{t}, \mathbf{u}_{D,t}\} = \argmin_{\substack{ 
        \vec{\mathcal{S}} \in \mathbb{R}^{n_1\times n_2} \\ 
        \mathbf{u}_{D} \in \mathbb{R}^{r_3}
    }} \Big\| \vec{\mathcal{P}}_t \circledast\Big( \vec{\mathcal{M}}_t - \mathcal{W}_t \times_3 \mathbf{u}_{D}^{\top} - \vec{\mathcal{S}}\Big)\Big\|_F^2 + \gamma \|\vec{\mathcal{S}}\|_1,
\end{equation}
\end{small}
where $ \mathcal{W}_t = \mathcal{G}_{t-1} \times_{1} \mathbf{U}_{T,t-1} \times_{2} \mathbf{U}_{S,t-1} $. Equation \ref{equ_5} is then reformulated into its matrix-vector form as follows:
\begin{equation}\label{equ_6}
    \{ \mathbf{s}_{t}, \mathbf{u}_{D,t}\} = \underset{\substack{  \mathbf{s} \in \mathbb{R}^{n_1 n_2 \times 1} \\ \mathbf{u}_{D} \in \mathbb{R}^{r_3 \times 1}}}{\operatorname{argmin}}\left\| \mathbf{P}_t \left(\mathbf{m}_t - {\mathcal{W}^{(3)}_t}^\top \mathbf{u}_{D}^\top -\mathbf{s}\right)\right\|_F^2 + \gamma \left\| \mathbf{s} \right\|_1,
\end{equation}
where $\mathbf{m}_t = \operatorname{vec}(\overrightarrow{\mathcal{M}_t}) $, $\mathbf{s}_t = \operatorname{vec}(\overrightarrow{\mathcal{S}_t})$, $\mathbf{P}_t = \operatorname{diag}(\operatorname{vec}(\overrightarrow{\mathcal{P}_t}))$. The optimal solutions for $\mathbf{s}_t$ and $\mathbf{u}_{D,t}$ in Equation \ref{equ_6} are obtained iteratively. Specifically, at the $i$-th iteration, we have:
\begin{equation}\label{equ_7}
    \mathbf{u}_{D,t,i} = \left({\mathcal{W}^{(3)}_t} \mathbf{P}_t {\mathcal{W}^{(3)}_t}^\top\right)^{-1} {\mathcal{W}^{(3)}_t} \mathbf{P}_t (\mathbf{m}_t - \mathbf{s}_{t,i-1}),
\end{equation}
\begin{equation}\label{equ_71}
    \mathbf{s}_{t,i} = \operatorname{Soft}_{\gamma}\left(\mathbf{P}_t(\mathbf{m}_t - {\mathcal{W}^{(3)}_t}^\top \mathbf{u}_{D,t,i}^\top)\right),
\end{equation}
where $ \operatorname{Soft}(\cdot) $ is the soft-thresholding operator for the $\ell_1 $-norm:
$$\operatorname{Soft}_{\gamma}(\cdot) := \operatorname{sign}(\cdot) \circ \max (|\cdot| - \gamma, 0),$$ and $ \gamma $ is a relaxation parameter. The iterative process is terminated when the residuals are sufficiently small, i.e., $  \operatorname{max}(\|\mathbf{s}_{t,i} - \mathbf{s}_{t,i-1}\|_F, \|\mathbf{u}_{t,i} - \mathbf{u}_{t,i-1}\|_F) \leq \epsilon $, or upon reaching the maximum number of iterations.
After detecting the sparse outlier \( \overrightarrow{\mathcal{S}_{t}} \), its impact on the data imputation is mitigated by the following outlier removal step:
$$
     \overrightarrow{\mathcal{P}_t} \circledast \overrightarrow{\mathcal{N}_t} = \overrightarrow{\mathcal{P}_t} \circledast (\overrightarrow{\mathcal{M}_t} - \overrightarrow{\mathcal{S}_t}).
$$

\subsection{Estimation of spatial factor matrix $\mathbf{U}_{S,t}$}

Based on the values for $\mathbf{u}_{D,t}$ and prior estimations for $\mathcal{G}_{t-1}$,$\mathbf{U}_{S,t-1}$, and  $\mathbf{U}_{T,t-1}$, the minimization problem outlined in Equation \ref{equ_3} for  $\mathbf{U}_{S,t}$  is reformulated as:
\begin{align}\label{equ_8}
    \mathbf{U}_{S,t} = \underset{\mathbf{U}_{S} \in \mathbb{R}^{n_1 \times r_1}}{\operatorname{argmin}}
    &\sum_{k=1}^t \lambda^{t-k} \Big\|\vec{\mathcal{P}}_k^{(2)} \circledast\Big(\vec{\mathcal{N}}_k^{(2)} - \mathbf{U}_{S} \mathcal{D}^{(2)}_k\Big)\Big\|_F^2 \nonumber \\
    &+ \alpha \operatorname{tr}\Big(\mathbf{U}_{S}^{\top} \mathbf{L} \mathbf{U}_{S}\Big),
\end{align}
where $\mathcal{D}_k$ is defined as:
$$
  \mathcal{D}_k= \mathcal{G}_{t-1} \times_{1} \mathbf{U}_{T,t-1}\times_3 {\mathbf{u}_{D,k}}^{\top}.  
$$

Indeed, the solution to Equation \ref{equ_8} involves decomposing the problem into subproblems, each targeting a specific row $\mathbf{U}_{S}[r,:]$ of the matrix $\mathbf{U}_{S}$, for $r = 1, 2, \ldots, n_2$. The optimization process for each row is encapsulated as follows:
\begin{align}\label{equ_10}
    \mathbf{U}_{S,t}[r,:] = &\mathop{\argmin}_{\mathbf{U}_{S}[r,:] \in \mathbb{R}^{r_1}} 
    \sum_{k=1}^t \lambda^{t-k}\Big\|\mathbf{P}_{k,r}^{(2)}\Big({\vec{\mathcal{N}}^{(2)}_{k}[r,:]}^{\top} \nonumber \\
    &\quad - {\mathcal{D}^{(2)}_k}^{\top}{\mathbf{U}_{S}[r,:]}^{\top}\Big)\Big\|_F^2 \nonumber + \alpha \mathbf{L}[r,r]\mathbf{U}_{S}[r,:]\mathbf{U}_{S}[r,:]^{\top} \nonumber \\
    &+ 2\alpha\sum_{c\neq r}\mathbf{L}[c,r]\mathbf{U}_{S}[c,:]\mathbf{U}_{S}[r,:]^{\top},
\end{align}
where $\mathbf{P}_{k, r}^{(2)}$ is defined as $\operatorname{diag}\left(\overrightarrow{\mathcal{P}}_k^{(2)}[r,:]\right)$.

To determine the optimal $\mathbf{U}_{S}[r,:]$, we set the derivative of Equation \ref{equ_10} to zero, yielding the following relationship:
\begin{small}
\begin{equation}
\begin{aligned}\label{equ_11}
\left(\sum_{k=1}^t \lambda^{t-k} \mathcal{D}^{(2)}_k \mathbf{P}_{k,r}^{(2)} {\mathcal{D}^{(2)}_k}^{\top} + \alpha \mathbf{L}[r,r]\mathbf{I}\right) {\mathbf{U}_{S,t}[r,:]}^{\top} \\= \sum_{k=1}^t \lambda^{t-k}\mathcal{D}^{(2)}_k \mathbf{P}_{k,r}^{(2)}{\overrightarrow{\mathcal{N}}^{(2)}_{k}[r,:]}^{\top}
- \alpha \sum_{c\neq r}\mathbf{L}[c,r]{\mathbf{U}_{S,t-1}[c,:]}^{\top}.
\end{aligned}
\end{equation}
\end{small}
Subsequently, Equation \ref{equ_11} simplifies to:
\begin{equation}\label{equ_12}
    \mathbf{R}_{S,r,t}{\mathbf{U}_{S,t}[r,:]}^{\top} = \mathbf{v}_{S,r,t},
\end{equation}
where
\begin{equation}\label{equ_13}
\begin{aligned}
    \mathbf{R}_{S,r,t} &= \sum_{k=1}^t \lambda^{t-k} {\mathcal{D}^{(2)}_k} \mathbf{P}_{r,k}^{(2)}  {\mathcal{D}^{(2)}_k}^{\top} + \alpha \mathbf{L}[r,r]\mathbf{I},\\\mathbf{v}_{S,r,t} &= \sum_{k=1}^t \lambda^{t-k}{\mathcal{D}^{(2)}_k}\mathbf{P}_{r,k}^{(2)}{\overrightarrow{\mathcal{N}}^{(2)}_{k}[r,:]}^{\top}\\
    &~~~~-\alpha \sum_{c\neq r}\mathbf{L}[c,r]{\mathbf{U}_{S,t-1}[c,:]}^{\top}.
\end{aligned}
\end{equation}


To facilitate Equation \ref{equ_11} update recursively, we subsequently redefine Equation \ref{equ_13} in the following manner:
\begin{equation}\label{equ_15}
\begin{aligned}
    \mathbf{R}_{S,r,t} &= \lambda \mathbf{R}_{S,r,t-1} + {\mathcal{D}^{(2)}_t} \mathbf{P}_{r,t}^{(2)}  {\mathcal{D}^{(2)}_t}^{\top} + \alpha(1-\lambda) \mathbf{L}[r,r]\mathbf{I},\\
    \mathbf{v}_{S,r,t} &=  \lambda \mathbf{v}_{S,r,t-1} + {\mathcal{D}^{(2)}_t} \mathbf{P}_{r,t}^{(2)}{\overrightarrow{\mathcal{N}}^{(2)}_{t}[r,:]}^{\top}- \alpha(1-\lambda) \\ &  \sum_{c\neq r}\mathbf{L}[c,r]{\mathbf{U}_{S,t-1}[c,:]}^{\top}.
\end{aligned}
\end{equation}

Upon integrating Equation \ref{equ_15} into Equation \ref{equ_12}, we derive:
\begin{align*}
    \mathbf{R}_{S,r,t}\mathbf{U}_{S,t}[r,:]^{\top} &= \mathbf{v}_{S,r,t} \\
    &= \lambda \mathbf{v}_{S,r,t-1} + \mathcal{D}^{(2)}_t \mathbf{P}_{r,t}^{(2)}\vec{\mathcal{N}}^{(2)}_{t}[r,:]^{\top} \\
    &\quad - \alpha(1-\lambda) \sum_{c\neq r}\mathbf{L}[c,r]\mathbf{U}_{S,t-1}[c,:]^{\top} \\
    &= \lambda \mathbf{R}_{S,r,t-1}\mathbf{U}_{S,t-1}[r,:]^{\top} \\
    &\quad + \mathcal{D}^{(2)}_t \mathbf{P}_{r,t}^{(2)}{\vec{\mathcal{N}}^{(2)}_{t}[r,:]}^{\top} \\
    &\quad - \alpha(1-\lambda) \sum_{c\neq r}\mathbf{L}[c,r]{\mathbf{U}_{S,t-1}[c,:]}^{\top} \\
    &= \Big(\mathbf{R}_{S,r,t} - \mathcal{D}^{(2)}_t \mathbf{P}_{r,t}^{(2)} {\mathcal{D}^{(2)}_t}^{\top} \\
    &\quad - \alpha(1-\lambda) \mathbf{L}[r,r]\mathbf{I}\Big){\mathbf{U}_{S,t-1}[r,:]}^{\top} \\
    &\quad + \mathcal{D}^{(2)}_t \mathbf{P}_{r,t}^{(2)}{\vec{\mathcal{N}}^{(2)}_{t}[r,:]}^{\top} \\
    &\quad - \alpha(1-\lambda) \sum_{c\neq r}\mathbf{L}[c,r]{\mathbf{U}_{S,t-1}[c,:]}^{\top},
\end{align*}
where $\mathbf{U}_{S,t-1}[r,:]$ denotes the previous time estimate for $r$-th row. This expression facilitates a parallel update across all rows of the spatial factor matrix $\mathbf{U}_{S,t}$, structured as follows:
\begin{align*}
    \mathbf{U}_{S,t}[r,:]^{\top} &= \mathbf{U}_{S,t-1}[r,:]^{\top}  + \mathbf{R}_{S,r,t}^{-1} \Big[ \mathcal{D}^{(2)}_t\mathbf{P}_{r,t}^{(2)} \\
    &\quad \times \Big(\vec{\mathcal{N}}^{(2)}_{t}[r,:]^{\top} - {\mathcal{D}^{(2)}_t}^{\top}\mathbf{U}_{S,t-1}[r,:]^{\top} \Big) \\
    &\quad - \alpha(1-\lambda) \sum_{c}\mathbf{L}[c,r]\mathbf{U}_{S,t-1}[c,:]^{\top} \Big].
\end{align*}
This recursive process efficiently updates each row by adjusting prior outputs with new tensor data and Laplacian regularization. Specifically, the update is given by:
$$\mathbf{P}_{r,t}^{(2)}\left({\overrightarrow{\mathcal{N}}^{(2)}_{t}[r,:]}^{\top}-{\mathcal{D}^{(2)}_t}^{\top}{\mathbf{U}_{S,t-1}}[r,:]^{\top} \right)\equiv \Delta {\overrightarrow{\mathcal{N}}_{t}^{(2)}[r,:]}^{\top},$$
where $\Delta\overrightarrow{\mathcal{N}_{t}}$ represents the residual error between the newly arrived tensor slice  at time $t$ and its reconstruction, which is defined as:
\begin{equation}\label{equ_res}
        \Delta \overrightarrow{\mathcal{N}}_t=\overrightarrow{\mathcal{P}}_t \circledast\left(\overrightarrow{\mathcal{N}_t}-\mathcal{W}_t \times_{3} {\mathbf{u}_{D,t}}^{\top}\right).
\end{equation}
Additionally, we define:
$$\mathcal{E}_{S,t}[r,:] = \sum_{c}\mathbf{L}[c,r]{\mathbf{U}_{S,t-1}}[c,:]^{\top}.$$
Ultimately, this iterative procedure culminates in the following update for each row of spatial matrix $\mathbf{U}_{S}$:
\begin{align}\label{equ_18}
    \mathbf{U}_{S,t}&[r,:]^{\top} = \mathbf{U}_{S,t-1}[r,:]^{\top} \nonumber \\
    &~~~~+ \mathbf{R}_{S,r,t}^{-1} \Bigl[ \mathcal{D}^{(2)}_t \Delta {\vec{\mathcal{N}}_{t}^{(2)}[r,:]}^{\top}- \alpha(1\lambda)\mathcal{E}_{S,t}[r,:]\Bigr].
\end{align}
\subsection{Estimation of temporal factor matrix $\mathbf{U}_{T,t}$}
In parallel to the adjustments made for $\mathbf{U}_{S,t}$, the minimization for $\mathbf{U}_{T,t}$ adopts a similar approach but focuses distinctly on different tensor unfoldings and constraints. Given the updates for $\mathbf{u}_{D,t}$ and the historical data from $\mathcal{G}_{t-1}$, $\mathbf{U}_{S,t-1}$, and $\mathbf{U}_{T,t-1}$, we redefine the optimization target for $\mathbf{U}_{T,t}$ as:
\begin{align}\label{equ_19}
        \mathbf{U}_{T,t}=\underset{\mathbf{U}_{T} \in \mathbb{R}^{n_1 \times r_1}}{\operatorname{argmin}} &\sum_{k=1}^t \lambda^{t-k} \left\|\overrightarrow{\mathcal{P}}_k^{(1)} \circledast\left(\overrightarrow{\mathcal{N}}_k^{(1)}-\mathbf{U}_{T} \mathcal{H}^{(1)}_k\right)\right\|_F^2 \nonumber \\  
        &+ \beta \left\| \mathcal{T}\mathbf{U}_{T}\right\|_F^2,
\end{align}
where $\mathcal{H}_k$ is defined as:

 $$\mathcal{H}_k= \mathcal{G}_{t-1} \times_{2} \mathbf{U}_{S,t-1} \times_3 \mathbf{u}_{D,k}^{\top}. $$

This differs from the mode-2 unfolding used in $\mathbf{U}_{S,t}$ by focusing on the spatial integration of subspace tracking. The decomposition strategy for solving Equation \ref{equ_19} mirrors that of Equation \ref{equ_8}, targeting each row $\mathbf{U}_{T,t}[r,:]$ within $\mathbf{U}_{T,t}$:
\begin{align*}
    \mathbf{U}_{T,t}[r,:] &= \argmin_{\mathbf{U}_{T}[r,:] \in \mathbb{R}^{r_1}} 
    \sum_{k=1}^t \lambda^{t-k} \Big\| \mathbf{P}_{r,k}^{(1)} \\
    &\quad \times \Big( {\vec{\mathcal{N}}_{k}^{(1)}[r,:]}^{\top} - {\mathcal{H}^{(1)}_k}^{\top}\mathbf{U}_{T}[r,:]^{\top} \Big) \Big\|_F^2 \\
    &+ \beta \| \mathbf{U}_{T}[r-1,:]^{\top} - \mathbf{U}_{T}[r,:]^{\top} \|_F^2 \\
    &+ \beta \| \mathbf{U}_{T}[r+1,:]^{\top} - \mathbf{U}_{T}[r,:]^{\top} \|_F^2,
\end{align*}
where the process for deriving the optimal $\mathbf{U}_{T,t}$ is analogous to that used in $\mathbf{U}_{S,t}$, but emphasizes temporal instead of spatial connections. Notably, when $r$ corresponds to the last row of $\mathbf{U}_{T}$, $r+1$ is interpreted as the first row of $\mathbf{U}_{T}$, thereby maintaining the cyclic structure of the temporal matrix. Therefore, we update each row of the temporal matrix $\mathbf{U}_{T}$ as follows:
\begin{align}\label{equ_29}
    \mathbf{U}_{T,t}&[r,:]^{\top} = \mathbf{U}_{T,t-1}[r,:]^{\top} \nonumber \\
    &~~+ \mathbf{R}_{T,r,t}^{-1} \Bigl[ \mathcal{H}_t \Delta\vec{\mathcal{N}}_{t}^{(1)}[r,:]^{\top} - \beta(1-\lambda)\mathcal{E}_{T,t}[r,:]\Bigr].
\end{align}
where 
\begin{align*}
    \mathbf{R}_{T,r,t} &= \lambda \mathbf{R}_{T,r,t-1} 
        + \mathcal{H}_t \mathbf{P}_{r,t}^{(1)} \mathcal{H}_t^{\top} + 2\beta(1-\lambda)\mathbf{I}, \\
    \mathcal{E}_{T,t}[r,:] &= 2\mathbf{U}_{T,t-1}[r,:]^{\top}  - \mathbf{U}_{T,t-1}[r-1,:]^{\top}\\
        &\quad - \mathbf{U}_{T,t-1}[r+1,:]^{\top}.
\end{align*}
This comprehensive update ensures the temporal subspace is accurately tracked and optimized across successive time steps.

\subsection{Estimation of core tensor $\mathcal{G}_t$}

In the context of updating the core tensor $\mathcal{G}_t$ with freshly revised loading factors, Equation \ref{equ_4} is reformulated as:
\begin{equation*} 
    \mathcal{G}_t=\underset{\mathcal{G}}{\operatorname{argmin}}\sum_{k=1}^t \lambda^{t-k}\left\|\overrightarrow{\mathcal{P}}_k^{(1)} \circledast\left(\overrightarrow{\mathcal{N}}_k^{(1)}-\mathbf{U}_{T,t} \mathcal{G}^{(1)} \mathbf{Z}_k^\top\right)\right\|_F^2,
\end{equation*}
where $\mathbf{Z}_k$ is constructed as:
\begin{equation*}
    \mathbf{Z}_k=\mathbf{u}_{D,k} \otimes \mathbf{U}_{S,t}.
\end{equation*}

Given the complexities associated with large-scale streaming data (i.e., significant $t$) and a high parameter count within $\mathcal{G}$ (i.e., large $\prod_{n=1}^N r_n$), the computational burden of traditional batch gradient methods may become prohibitive. To enhance computational efficiency, a stochastic approximation approach \cite{abed2023tracking} is suggested:
\begin{equation*}
    \mathcal{G}_t=\underset{\mathcal{G}}{\operatorname{argmin}}\left\|\overrightarrow{\mathcal{P}}_k^{(1)} \circledast\left(\overrightarrow{\mathcal{N}}_k^{(1)}-\mathbf{U}_{T,t} \mathbf{G}^{(1)} \mathbf{Z}_k^\top\right)\right\|_F^2.
\end{equation*}

This method focuses on minimizing the error associated with the most recent tensor slice, thereby aligning the core tensor closely with the latest observations. Considering that $\Delta \overrightarrow{\mathcal{N}}_t^{(1)},$ 
derived from Equation \ref{equ_res}, is equivalent to $\mathbf{P}_t^{(1)} \circledast\left(\overrightarrow{\mathcal{N}}_t^{(1)}-\mathbf{U}_{T,t} \mathbf{G}_{t-1}^{(1)} \mathbf{Z}_t^\top\right)$, the update for $\mathcal{G}$ at time $t$ can be derived as follows:
\begin{equation}\label{equ_34}
    \Delta \overrightarrow{\mathcal{N}}_t^{(1)}=\overrightarrow{\mathcal{P}}_t^{(1)} \circledast\left(\mathbf{U}_{T,t} \Delta \mathcal{G}_t^{(1)} \mathbf{Z}_t^\top\right),
\end{equation}
where $\Delta \mathcal{G}_t^{(1)}=\mathcal{G}_t^{(1)}- \mathcal{G}_{t-1}^{(1)}$. Equation \ref{equ_34} leads to a straightforward calculation for the change in the core tensor:
\begin{equation*}
   \Delta \mathcal{G}_t^{(1)}=\left(\mathbf{U}_{T,t}\right)^{\dagger} \Delta \overrightarrow{\mathcal{N}}_t^{(1)} {\mathbf{Z}_t^\top}^{\dagger} . 
\end{equation*}
This incremental adjustment $\Delta \mathcal{G}_t^{(1)}$ is subsequently reshaped into the three-dimensional tensor $\Delta \mathcal{G}_t$, paving the way for a simple yet effective update rule:
\begin{equation}\label{equ_36}
    \mathcal{G}_t=\mathcal{G}_{t-1}+\Delta \mathcal{G}_t.
\end{equation}

Overall, the scheme for solving (\ref{equ_3}) is summarized in Algorithm \ref{STORTD}.
\begin{algorithm}[H]
	\caption{Spatio-Temporal Online Robust Tucker Decomposition  (STORTD)}
	\label{STORTD}
	\begin{algorithmic}[1]
            \renewcommand{\algorithmicrequire}{ \textbf{Input:}}
		\REQUIRE Sequentially observed data $\mathcal{M} = \overrightarrow{\mathcal{M}_{1}}\boxplus \overrightarrow{\mathcal{M}_{2}} \boxplus \cdots \boxplus \overrightarrow{\mathcal{M}_{T}} \in \mathbb{R}^{n_1 \times n_2 \times T}$, forgetting factor $\lambda \in [0,1]$, Tucker rank $\mathbf{r}= [r_1,r_2,r_3]$, regularization parameters $\alpha$, $\beta$.
		\FOR{$t=1,2,\ldots,T$}
            \STATE  Update sparse tensor slice $\overrightarrow{\mathcal{S}_{t}} $ by equation (\ref{equ_71});
            \STATE  Update weight $\mathbf{u}_{D,t}$ by equation (\ref{equ_7});
		\STATE  Update spatial factor matrix $\mathbf{U}_{S,t} $ by equation (\ref{equ_18});
  \STATE  Update temporal factor matrix $\mathbf{U}_{T,t}$ by equation (\ref{equ_29});
		\STATE  Update core tensor $\mathcal{G}_{t} $ by equation (\ref{equ_36});
  	    \STATE  Estimated  low-rank tensor slice $\overrightarrow{\mathcal{N}_{t}} \leftarrow \mathcal{G}_{t} \times_{1} \mathbf{U}_{T,t} \times_{2} \mathbf{U}_{S,t} \times_3 {\mathbf{u}_{D,t}}^{\top}$;
            \STATE $\mathcal{N}[:,:,t]\leftarrow \overrightarrow{\mathcal{N}_{t}}$, $\mathcal{S}[:,:,t]\leftarrow \overrightarrow{\mathcal{S}_{t}}$.
            \ENDFOR

        \renewcommand{\algorithmicrequire}{ \textbf{Output:}}
        \REQUIRE Estimated  low-rank tensor $\mathcal{N}$ and outlier tensor $\mathcal{S}$. 
        
	\end{algorithmic}
\end{algorithm}

\subsection{Complexity and storage cost}
For our analysis, we consider a streaming tensor with cubic dimensions $n \times n \times n$, where each frontal slice also measures $n \times n $. We define the tensor rank as $[r,r,r]$, and given that $r$ is substantially smaller than $n$, it is assumed $r^2<n$.

The computational load of the STORTD method primarily hinges on four key computations: (1) calculating the weight vector $\mathbf{u}_t$ , (2) determining the spatial tensor factors $\mathbf{U}_{S}$, (3) computing the temporal tensor factors $\mathbf{U}_{T}$, and (4) deriving the core tensor $\mathcal{G}$. The first calculation, as well as the computation of both the spatial and temporal factors, incurs a complexity of $\mathcal{O}\left(\left|\Omega_t\right| r^2\right)$, primarily due to matrix multiplication. The computation of $\Delta \mathcal{G}$ in the final estimation requires $\mathcal{O}\left(n^2 r+n r^4\right)$, where $n^2 r$ arises from the matrix multiplication and $n r^4$ is due to the computation of the pseudoinverse.

Regarding memory requirements, STORTD necessitates $\mathcal{O}\left(r^3\right)$ and $\mathcal{O}\left(nr\right)$ words to store the core tensor $\mathcal{G}$ and the matrix factors  $\mathbf{U}_{S}$ and $\mathbf{U}_{T}$, respectively. Additionally, storing the matrix $\mathbf{R}_{S,r,t}$ and $\mathbf{R}_{T,r,t}$ consume $\mathcal{O}\left(r^2\right)$ words of memory, while the storage for the Laplacian matrix 
$\mathbf{L}$ and the Tepolize matrix $\mathcal{T}$ each require $\mathcal{O}\left(n^2\right)$ words. In total, maintaining STORTD's data structure at each timestep $t$ demands $\mathcal{O}\left(n^2 \right)$ words of memory. Consequently, the memory requirements remain constant regardless of the sample size, which fulfills the requirements of large-scale ITS.

\section{Experiment}
In this section, relying on transportation domain datasets, we experimentally investigate the performance of the STORTD models under the different missing scenarios.

\subsection{Experimental Settings}

In our empirical analysis, we utilize three distinct datasets to evaluate the performance of the proposed models: passenger flow data from the Hangzhou metro system\footnote{\url{https://tianchi.aliyun.com/competition/entrance/231708/information}}, traffic speed data from Guangzhou\footnote{\url{http://www.openits.cn/openData2/792.jhtml}}, and traffic flow data from San Bernardino, captured by the PeMS network\footnote{\url{https://pems.dot.ca.gov/}}. The Hangzhou metro dataset encompasses observations from 80 stations over a period from January 1, 2019, to January 25, 2019, totaling 25 days. For each station, passenger flow data are recorded in 10-minute intervals, resulting in 108 observations per day across 18 hours of metro operation. This dataset is structured as a third-order tensor of dimensions $108 \times 80 \times 25$, representing time, stations, and days, respectively. Similarly, the Guangzhou traffic dataset, which spans from August 1, 2016, to September 30, 2016 (61 days), includes speed data from 214 road segments that primarily consist of urban expressways and arterial roads. Data for each segment is also recorded in 10-minute intervals, providing 144 speed observations per day. This dataset forms a $144 \times 214 \times 61$ third-order tensor. Lastly, the PeMS dataset, covering the period from July to August 2016, involves data from 170 detectors located across 8 roads, lasting for 62 days. Traffic flow data is collected every 5 minutes, yielding 288 daily observations for each detector. Consequently, this data can be modeled as a $288 \times 170 \times 62$ third-order tensor. 

\begin{figure}[htbp]
  \centering
  \begin{subfigure}[t]{0.15\textwidth}
    \centering
    \includegraphics[width=\linewidth, scale=1]{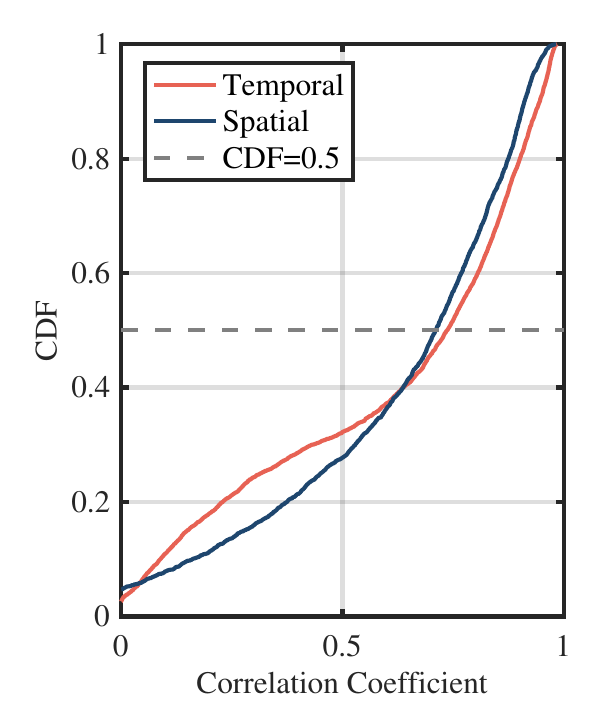} 
    \caption{Hangzhou}\label{ds:sub1}
  \end{subfigure}
  \begin{subfigure}[t]{0.15\textwidth}
    \centering
    \includegraphics[width=\linewidth, scale=1]{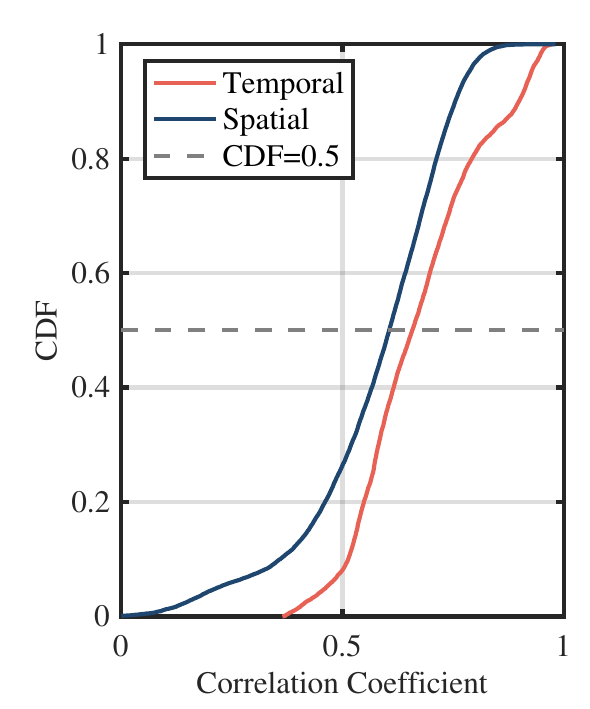}
    \caption{Guangzhou}\label{ds:sub2}
  \end{subfigure}
  \begin{subfigure}[t]{0.15\textwidth}
    \centering
    \includegraphics[width=\linewidth, scale=1]{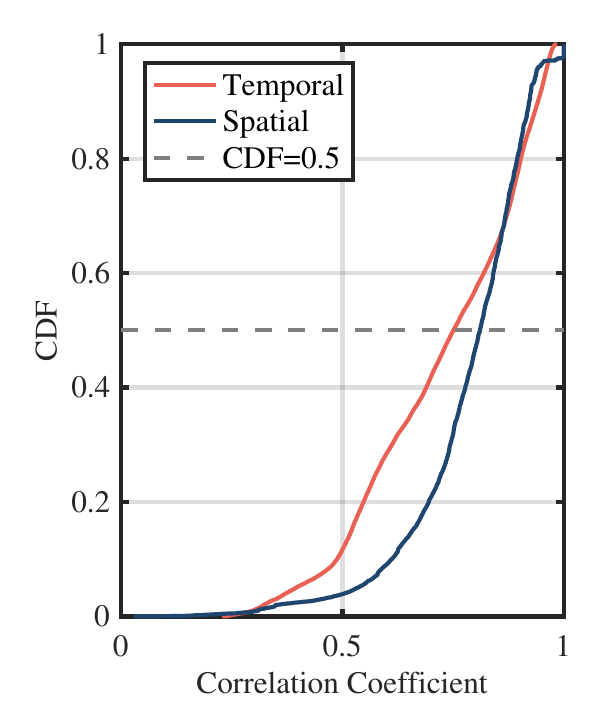}
    \caption{PeMS}\label{ds:sub3}
  \end{subfigure}  
  \caption{Cumulative distribution function}\label{ds}
\end{figure}

To demonstrate the necessity of our proposed algorithm in the context of traffic data recovery, we begin by analyzing the spatio-temporal characteristics of real-world traffic datasets. Specifically, denoting the entire dataset as $\mathcal{N}^{*}$, we compute spatial correlations between pairs of rows in ${\mathcal{N}^{*}}^{(1)}$ and temporal correlations between rows in ${\mathcal{N}^{*}}^{(2)}$. The cumulative distribution functions (CDFs) of the resulting correlation coefficients are plotted in Figure \ref{ds}. The results show that more than half of the nodes exhibit strong correlations in both the spatial and temporal dimensions, revealing clear spatio-temporal dependencies across all the three datasets.

In our experiments, we randomly masked 20\% to 80\% of the data entries as missing and applied both the proposed method and baseline algorithms to perform imputation.  As outlined in Section \ref{section: Problem Definition}, we have devised four missing patterns: random missing (RM, see Figure \ref{mp:sub1}), temporal missing (TM, Figure \ref{mp:sub2}), spatial missing (SM, Figure \ref{mp:sub3}) and mixed missing (MM, Figure \ref{mp:sub4}). Among these, the TM, SM and MM scenarios present greater challenges than the RM scenario, as the missing data are corrupted in a correlated manner. These four missing scenarios can help us to comprehensively evaluate the performance and effectiveness of different models. The observed entries are used for training, while the masked entries serve as the test set. To evaluate model performance, we adopt the Relative Squared Error (RSE) metric, defined as:
$$
\operatorname{RSE} = \sqrt{\frac{\sum_{i=1}^n (x_i - \hat{x}_i)^2}{\sum_{i=1}^n x_i^2}},
$$
where $x_i$ and $\hat{x}_i$ denote the ground truth and the recovered values, respectively, and $n$ is the total number of test entries.

\subsection{Parameter Sensitivity Experiment}

\begin{figure}[H]
  \centering

  \begin{subfigure}[b]{0.22\textwidth}
    \centering
    \includegraphics[width=\linewidth, scale=1]{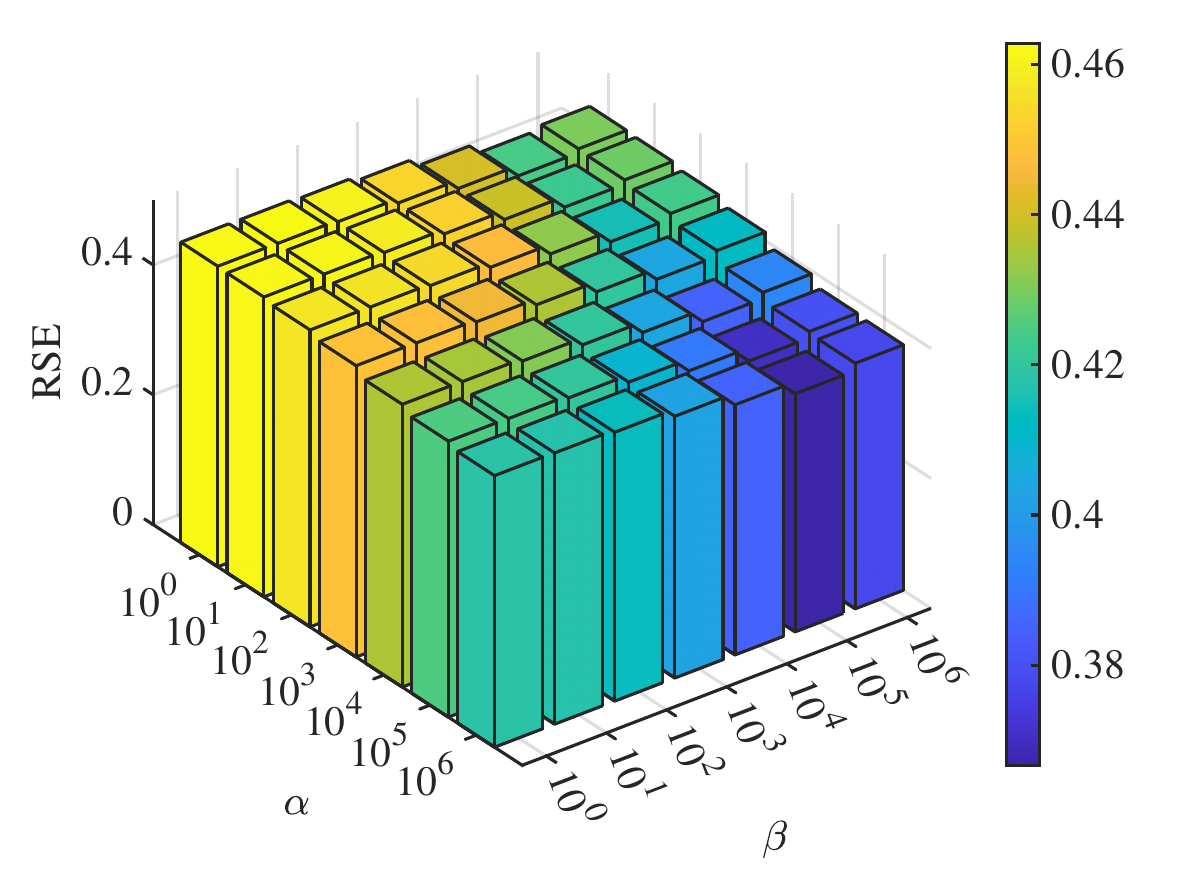}
    \caption{RM}
    \label{ps:sub1}
  \end{subfigure}
  \hfill 
  \begin{subfigure}[b]{0.22\textwidth}
    \centering
    \includegraphics[width=\linewidth, scale=1]{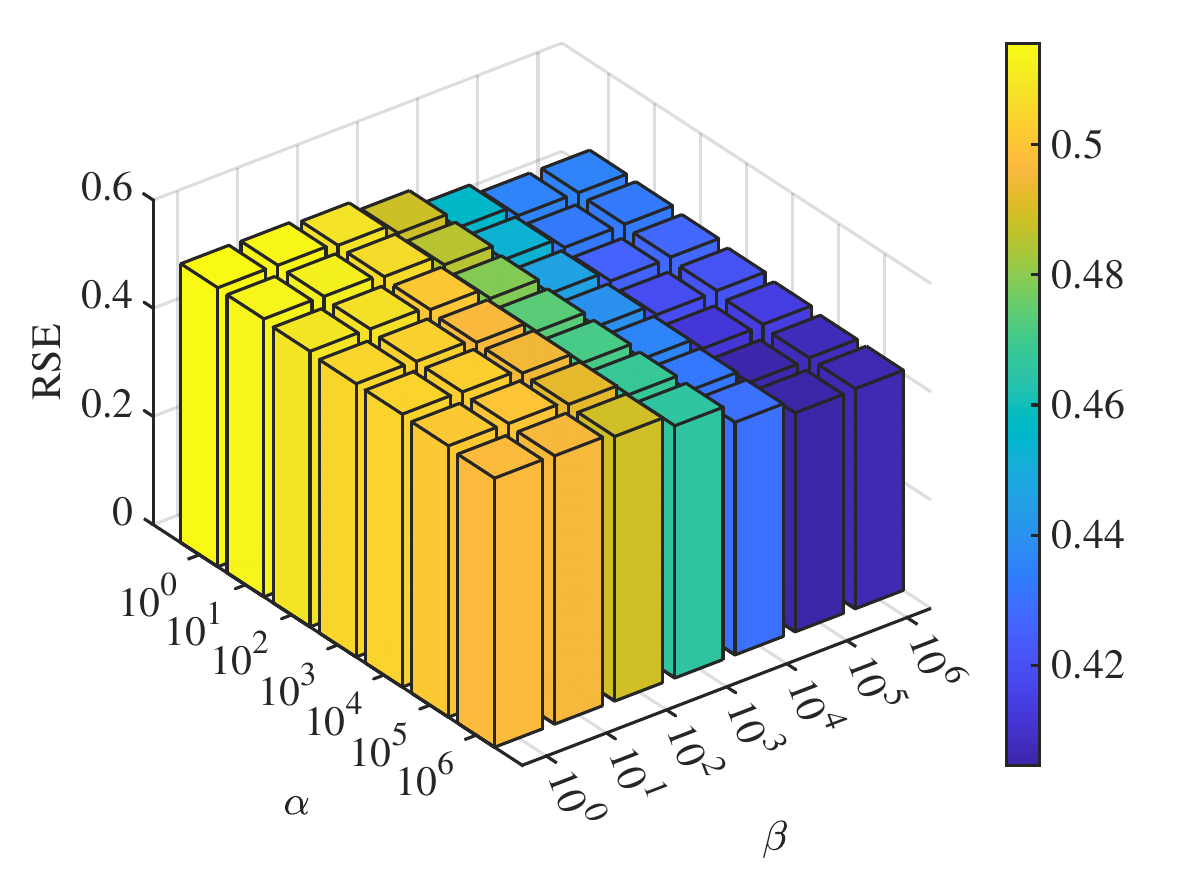}
    \caption{TM}
    \label{ps:sub2}
  \end{subfigure}

  \begin{subfigure}[b]{0.22\textwidth}
    \centering
    \includegraphics[width=\linewidth, scale=1]{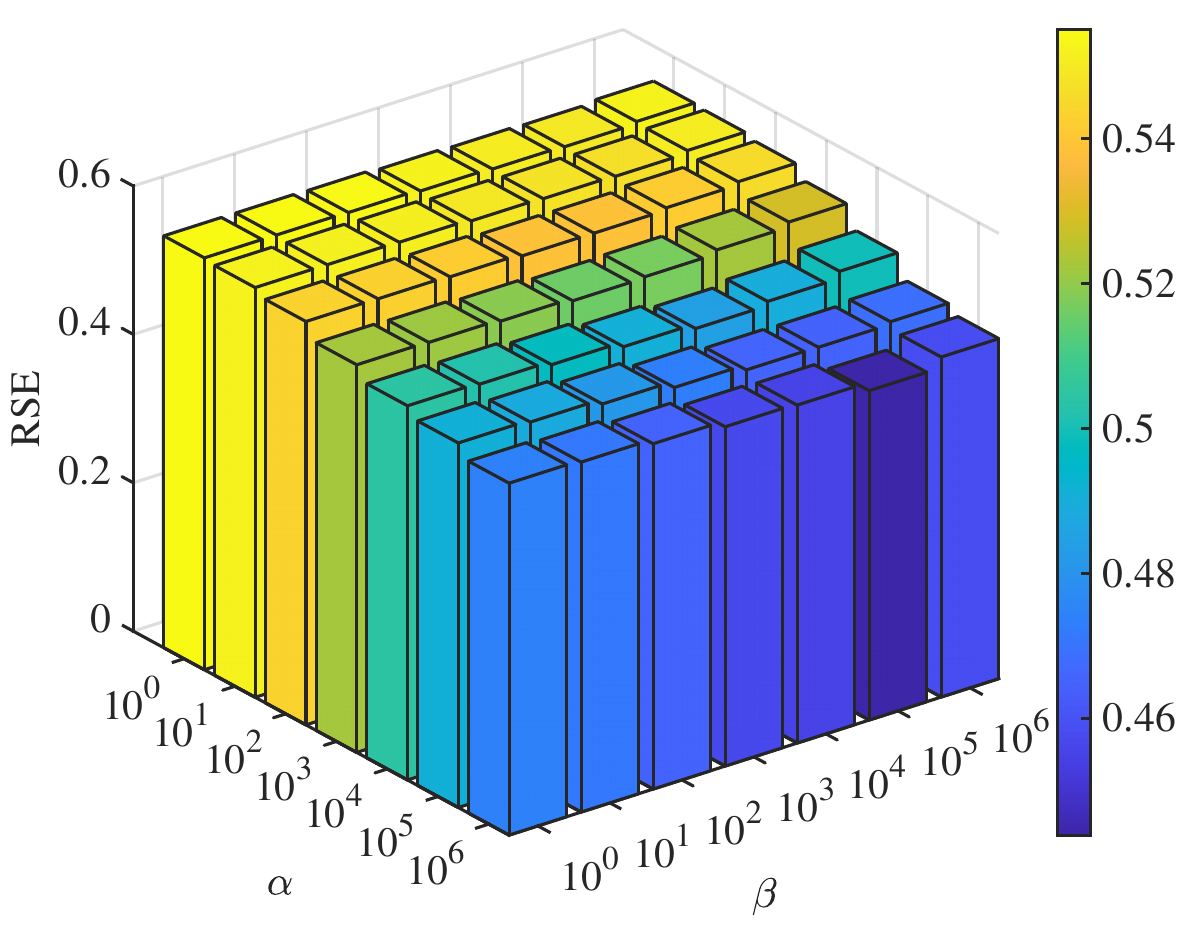}
    \caption{SM}
    \label{ps:sub3}
  \end{subfigure}
  \hfill 
  \begin{subfigure}[b]{0.22\textwidth}
    \centering
    \includegraphics[width=\linewidth, scale=1]{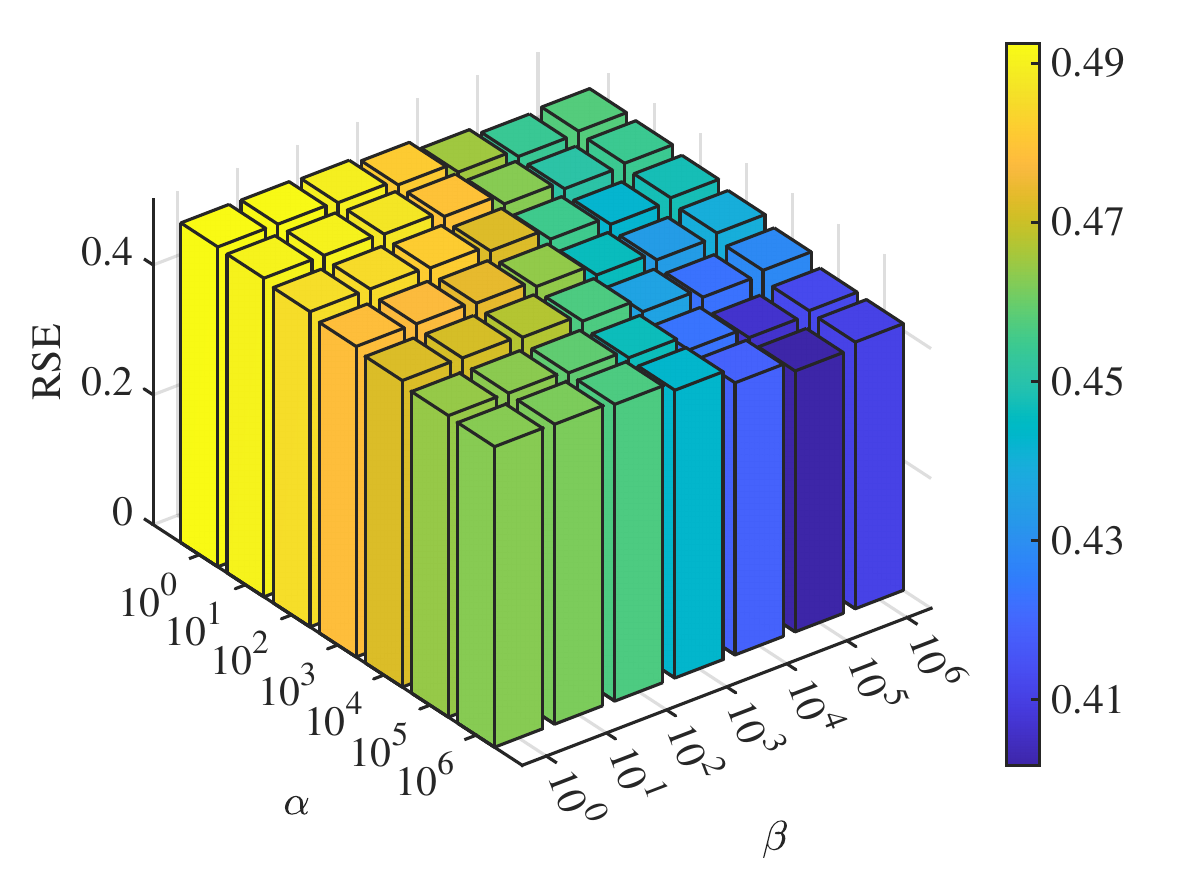}
    \caption{MM}
    \label{ps:sub4}
  \end{subfigure}
  
  \caption{Recovery performance under different values of $\alpha$ and $\beta$ across four missing scenarios}
  \label{ps}
\end{figure}

In this section, we conduct a parameter sensitivity experiment on the Hangzhou dataset using our STORTD model, focusing on the hyper-parameters $\alpha$ and $\beta$ which modulate the intensity of spatial and temporal regularization, respectively. Both parameters were varied across a logarithmic scale $\{10^0, 10^1, 10^2, 10^3, 10^4, 10^5, 10^6\}$ to explore their effects under various missing data scenarios. This experimental procedure was iterated ten times to ensure reliability, with the average results illustrated in Figure \ref{ps}.

The results presented in Figure~\ref{ps} lead to the following key observations:
\begin{enumerate}
    \item \textbf{Spatial Missing (SM)}: The algorithm demonstrates high sensitivity to $\alpha$. Larger values of $\alpha$ enforce stronger spatial regularization, which significantly improves recovery performance under spatially missing data.
    
    \item \textbf{Temporal Missing (TM)}: Similarly, the effect of $\beta$ highlights its importance in temporal regularization. Increasing $\beta$ leads to more effective recovery in cases with temporal missing patterns.
    
    \item \textbf{Random Missing (RM) and Mixed Missing (MM)}: The combined impact of spatial and temporal constraints becomes evident. Particularly, $\alpha$ and $\beta$ values within the range of $10^4$ to $10^6$ achieve markedly better performance than lower settings (e.g., $10^0$–$10^2$), suggesting the necessity of balanced regularization in complex missing scenarios.
\end{enumerate}

These findings highlight the critical role of parameter tuning in enhancing tensor recovery performance for ITS, ultimately contributing to more reliable traffic modeling and prediction.

\subsection{Ablation Experiment}
\begin{figure}[htbp]
  \centering
  \captionsetup[subfigure]{font=footnotesize}
  \begin{subfigure}[b]{0.45\textwidth}
    \centering
    \includegraphics[width=\linewidth, height=0.12\textheight]{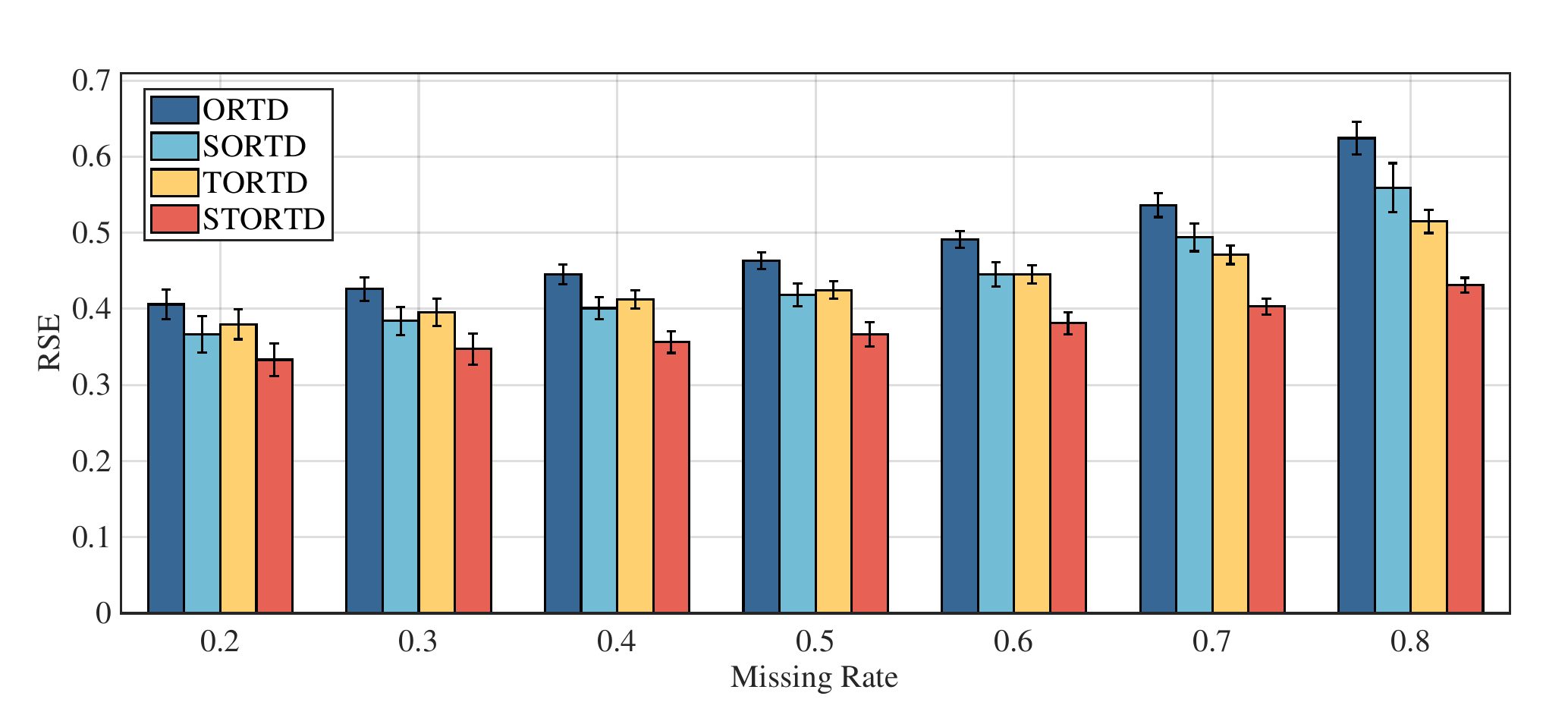}
    \caption{RM}\label{abl:sub1}
  \end{subfigure}
  
  
  \begin{subfigure}[b]{0.45\textwidth}
    \centering
    \includegraphics[width=\linewidth, height=0.12\textheight]{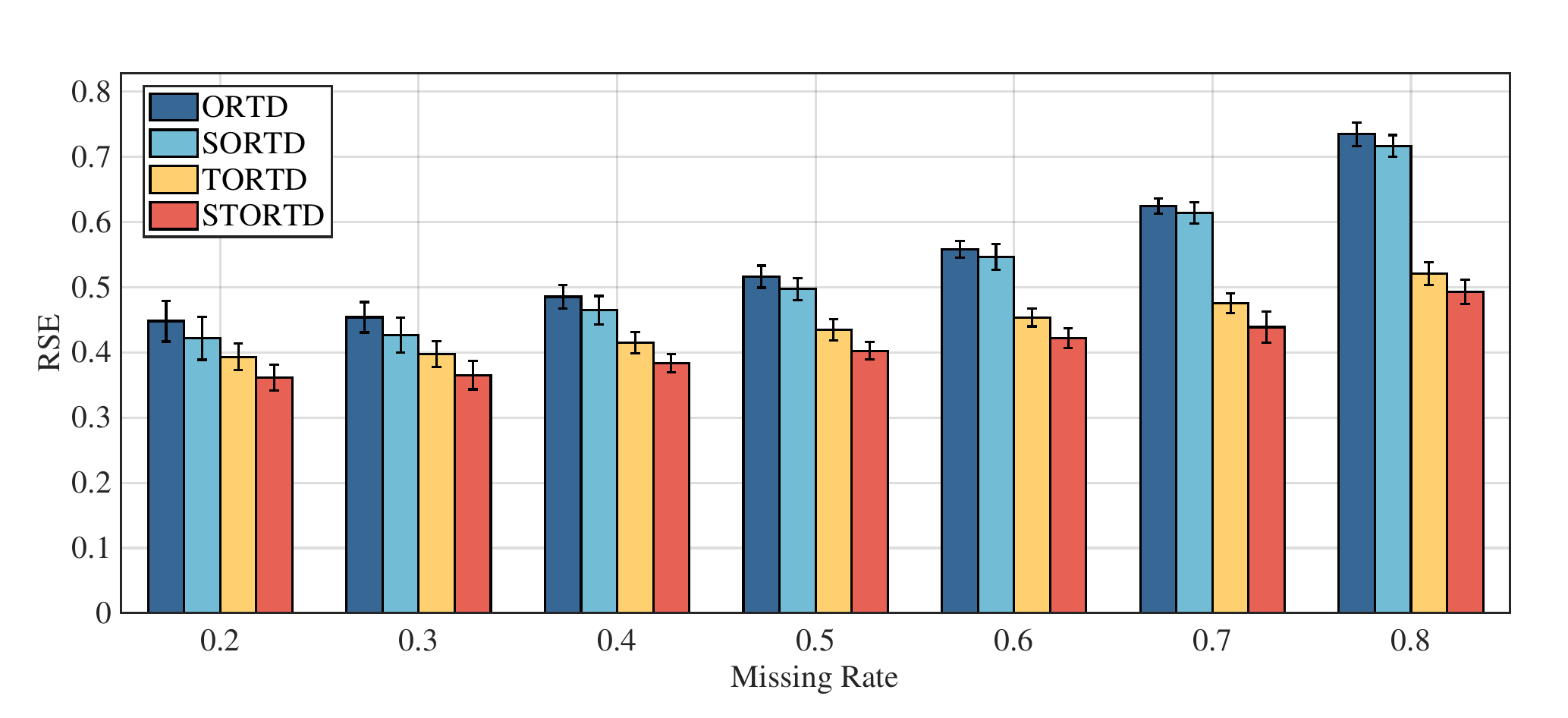}
    \caption{TM}\label{abl:sub2}
  \end{subfigure}
  
  
  \begin{subfigure}[b]{0.45\textwidth}
    \centering
    \includegraphics[width=\linewidth, height=0.12\textheight]{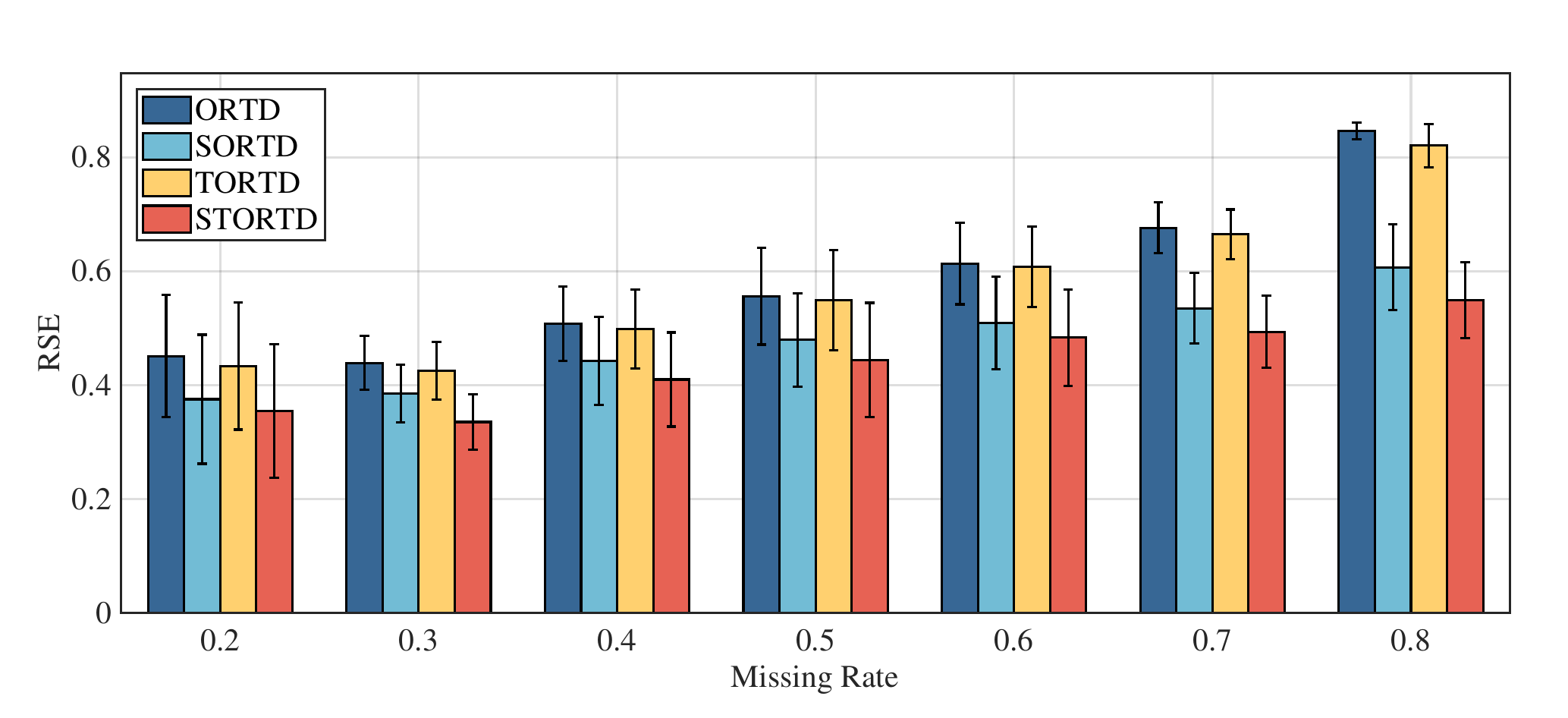}
    \caption{SM}\label{abl:sub3}
  \end{subfigure}
  
  
  \begin{subfigure}[b]{0.45\textwidth}
    \centering
    \includegraphics[width=\linewidth, height=0.12\textheight]{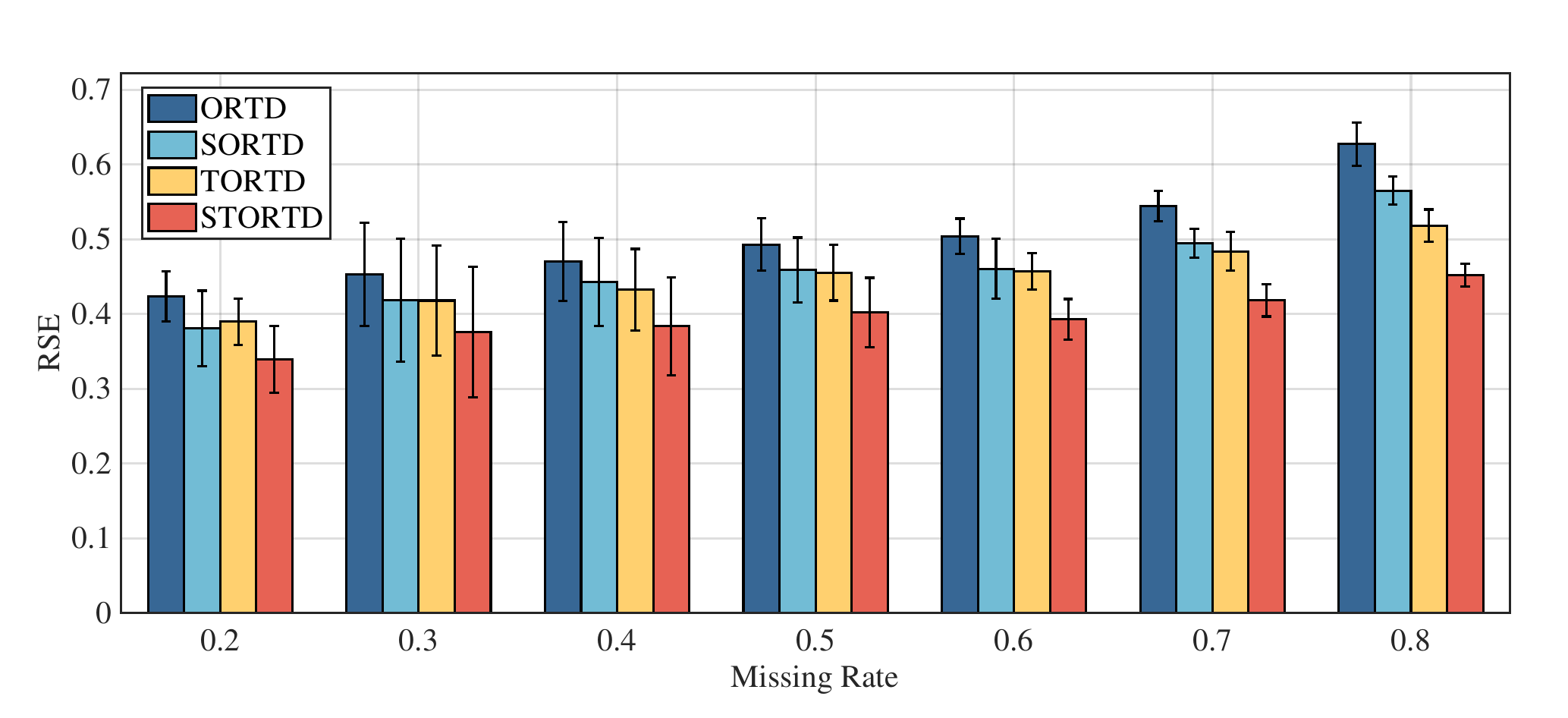}
    \caption{MM}\label{abl:sub4}
  \end{subfigure}
  \caption{The recovery results due to different ablation methods in four scenarios}\label{abl}
\end{figure}

In this section, we conduct a parameter ablation study on the Hangzhou dataset to evaluate the individual and joint effects of spatial and temporal regularization. Specifically:
\begin{itemize}
    \item ORTD: We set both $\alpha = 0$ and $\beta = 0$, corresponding to an online robust Tucker decomposition without any spatio-temporal regularization.
    \item SORTD: We fix $\alpha = 0$ and select $\beta$ over the set $\{10^0, 10^1, \dots, 10^6\}$, representing a model with only temporal regularization.
    \item TORTD: We fix $\beta = 0$ and select $\alpha$ over the same set $\{10^0, 10^1, \dots, 10^6\}$, representing a model with only spatial regularization.
    \item STORTD: Both $\alpha$ and $\beta$ are selected from the set $\{10^0, 10^1, \dots, 10^6\}$, thereby modeling joint spatial-temporal regularization.
\end{itemize}
We evaluate recovery performance under various missing patterns. Each experiment is repeated ten times, and the averaged results are reported in Figure~\ref{abl}.

From Figure \ref{abl}, we can observe that: 
\begin{enumerate}
\item ORTD, which does not incorporate any regularization, consistently underperforms compared to the three regularized variants across all missing scenarios. This suggests that exploiting local spatio-temporal consistency provides a valuable structural complement to global low-rank priors, thereby consistently enhancing traffic data recovery performance. 

\item SORTD and TORTD show greater advantages under the SM and TM missing patterns, respectively, indicating that incorporating spatial or temporal regularization is particularly beneficial when the missingness is concentrated along the corresponding dimension. 

\item STORTD outperforms all other variants across different missing rates and patterns, confirming that simultaneously leveraging both spatial and temporal local consistencies yields superior and more stable recovery results.
\end{enumerate}

\subsection{Comparative Experiment}
In the comparative experiments, we evaluate the proposed STORTD against several widely used data imputation methods. All methods are implemented in MATLAB 2021a on a computer with an Intel Xeon Gold 5120 2.20GHz CPU. Each method is run ten times, and the average results are reported.

\subsubsection{Compared with Online Algorithm}
In this section, we compare the proposed STORTD with three online matrix completion algorithms, namely, GRASTA\cite{he2011online}, OLRSC\cite{shen2016online},PETRELS-ADMM\cite{dung2021robust}, as well as six online tensor completion algorithms, including OLRTR\cite{hu2022streaming},OLSTEC\cite{kasai2019fast}, OSTD\cite{sobral2015online}, TeCPSGD\cite{mardani2015subspace}, OLRTSC\cite{wu2022online} and BSTF\cite{zhang2018variational}. The evaluation is conducted on three widely used traffic datasets: Hangzhou, Guangzhou, and PeMS, as introduced earlier.

\begin{figure*}[t]
  \centering
  \captionsetup[subfigure]{font=footnotesize}
  
  \begin{subfigure}[b]{0.4\linewidth}
    \centering
    \includegraphics[width=\linewidth]{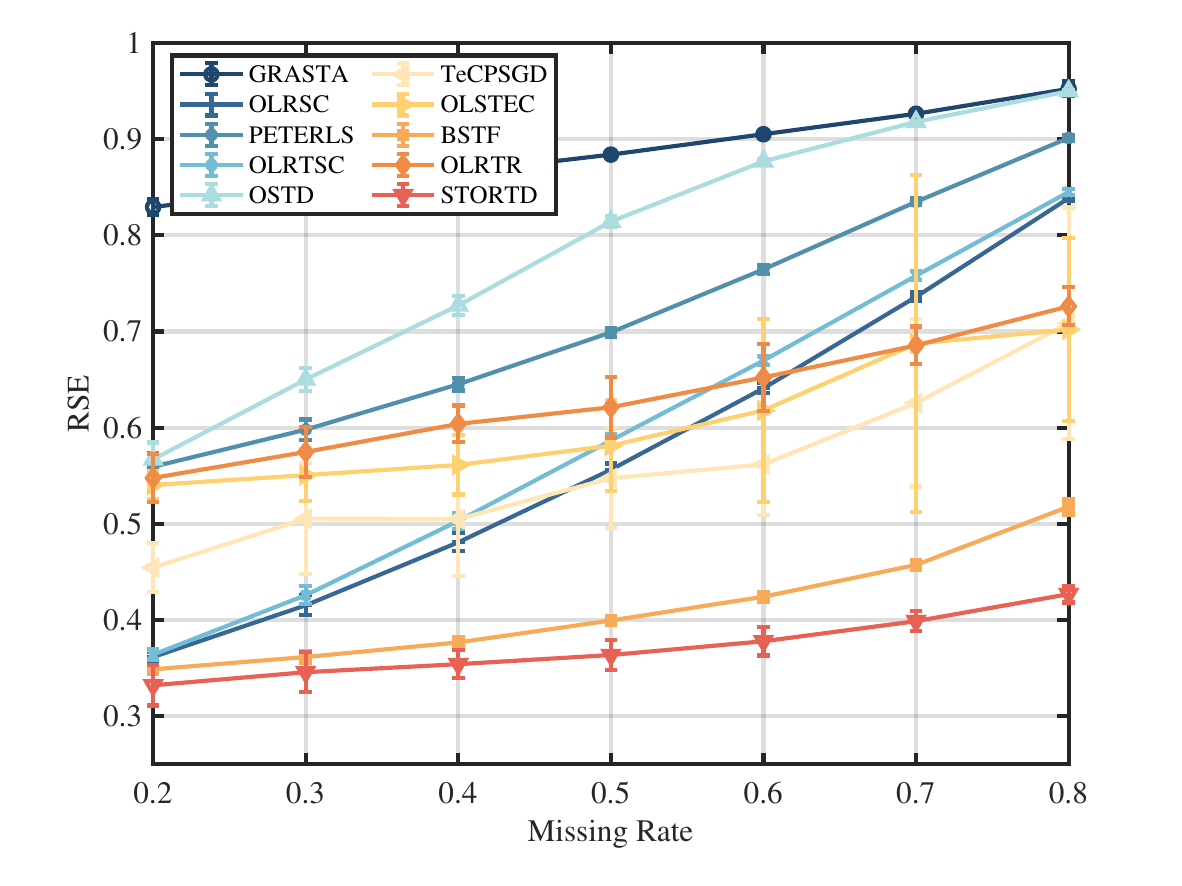}
    \caption{RM}\label{hzcp:sub1}
  \end{subfigure}
  \begin{subfigure}[b]{0.4\linewidth}
    \centering
    \includegraphics[width=\linewidth]{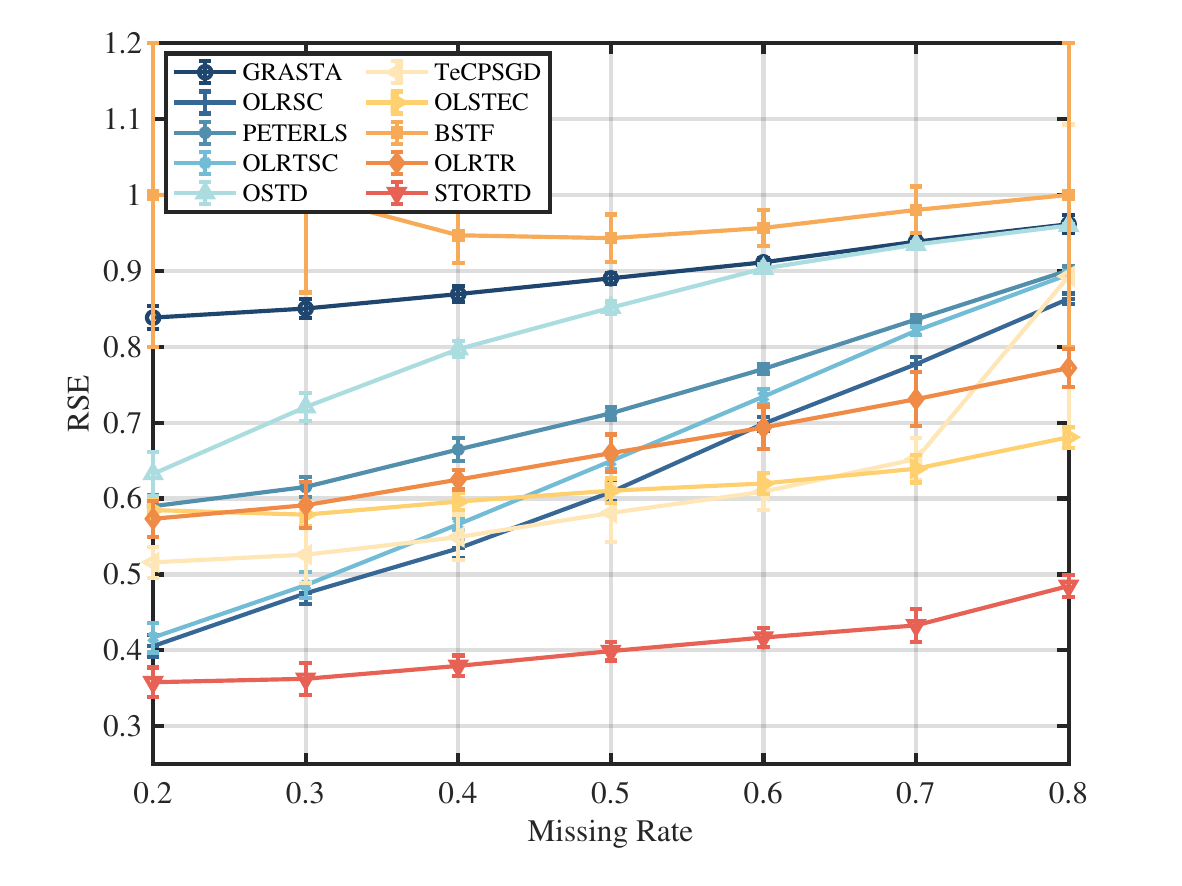}
    \caption{TM}\label{hzcp:sub2}
  \end{subfigure}

  \begin{subfigure}[b]{0.4\linewidth}
    \centering
    \includegraphics[width=\linewidth]{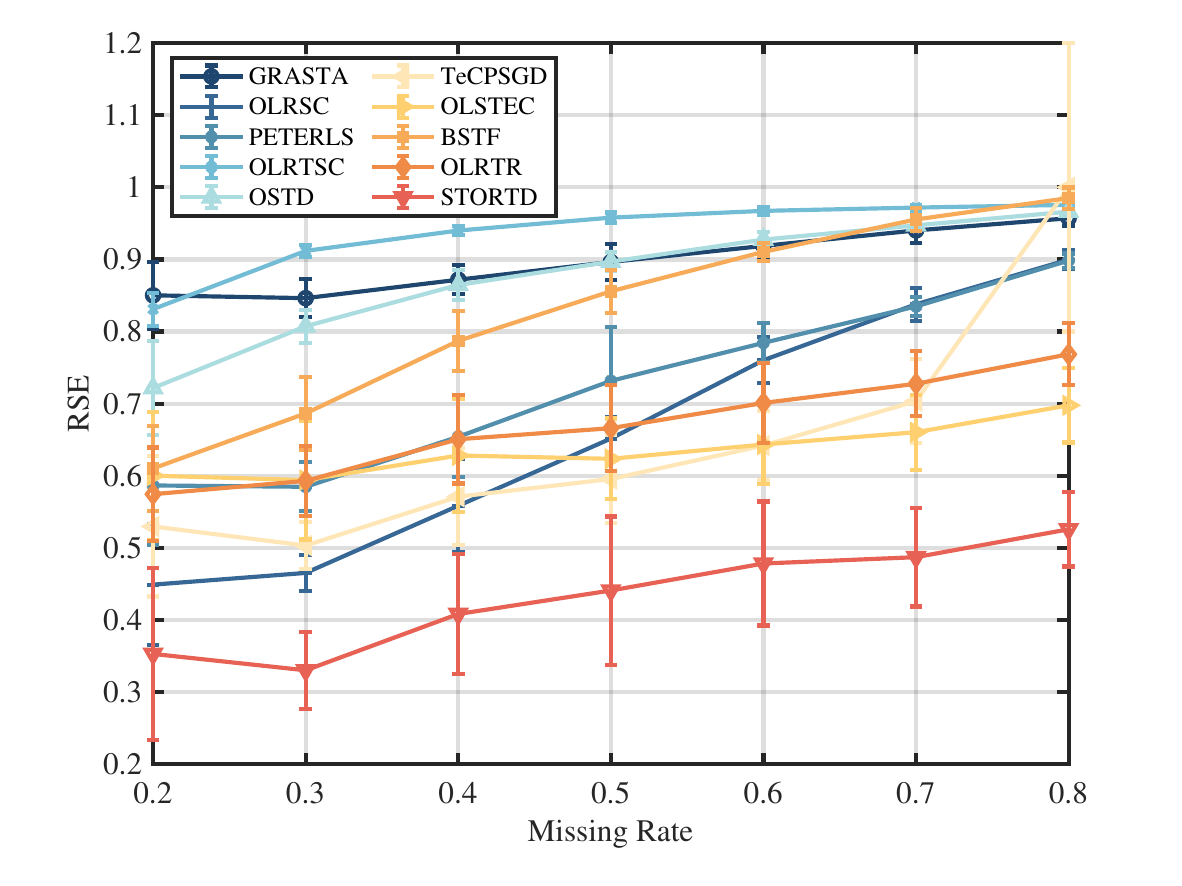}
    \caption{SM}\label{hzcp:sub3}
  \end{subfigure}
  \begin{subfigure}[b]{0.4\linewidth}
    \centering
    \includegraphics[width=\linewidth]{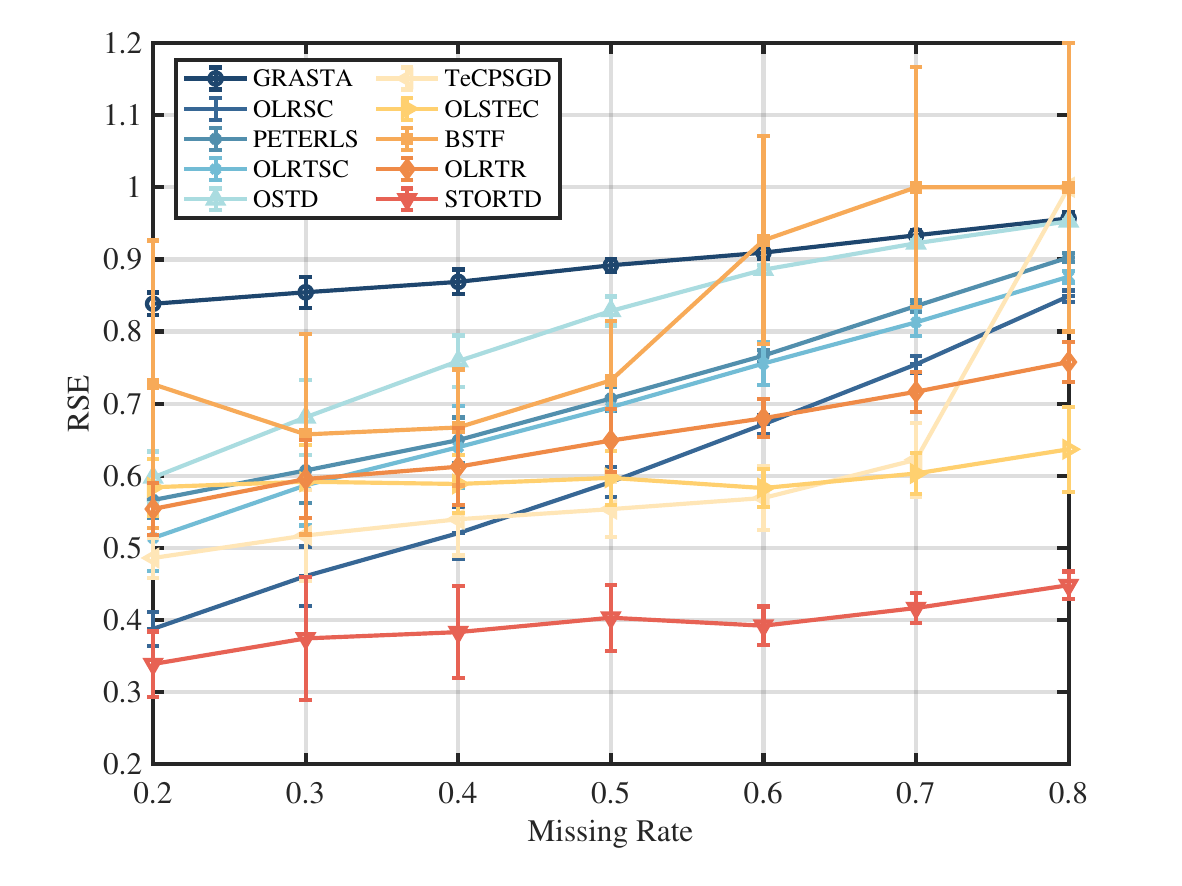}
    \caption{MM}\label{hzcp:sub4}
  \end{subfigure}
  
  \caption{Comparison with online algorithms on the Hangzhou dataset}
  \label{hzcp}
\end{figure*}

The recovery results on the Hangzhou dataset are presented in Figure~\ref{hzcp}, while the results on the Guangzhou and PeMS datasets are provided in Fig.1 and Fig.2 in the supplementary material due to page limitations. Based on these results, we make the following observations:
\begin{enumerate}
\item Overall, the proposed STORTD consistently outperforms all compared algorithms in terms of recovery accuracy. This advantage mainly stems from its joint exploitation of spatial and temporal local consistency, which enables more accurate imputation of streaming traffic data.

\item As the missing ratio increases, the recovery performance of all compared algorithms deteriorates. In particular, OLRSC and PETRELS exhibit substantial performance degradation at 80\% missingness, with their RSE values more than doubling compared to those at 20\%. In contrast, our method maintains strong robustness, with RSE increasing by less than 40\% over the same range. 

\item Compared to the random missing scenario (RM), the performance gap between our algorithm and the baselines becomes more pronounced under structured missing scenarios (i.e., SM, TM, and MM), indicating that the incorporation of spatio-temporal constraints makes our method particularly effective for real-world traffic data recovery where missingness often follows structured patterns.
\end{enumerate}

In summary, STORTD demonstrates remarkable robustness and generalization capabilities, consistently maintaining stable recovery performance even under extreme missing rates and complex missing patterns. Its adaptability and reliability make it well-suited for deployment in complex, real-world ITS environments.

\subsubsection{Compared with Offline Algorithm}
We then compare the proposed STORTD with nine representative batch-based completion algorithms, including SNN\cite{lu2018unified}, TNN\cite{lu2018exact}, LRSTD-PALM\cite{gongspatiotemporal}, LRSTD-IALM\cite{gongspatiotemporal}, BATF\cite{chen2019missing}, BGCP\cite{chen2019bayesian}, HaLRTC\cite{liu2012tensor}, LRTC-TNN\cite{chen2020nonconvex}, and LCR-2D\cite{chen2024laplacian}.

\begin{figure*}[htbp]
  \centering
  \includegraphics[width=0.8\linewidth,scale=0.8]{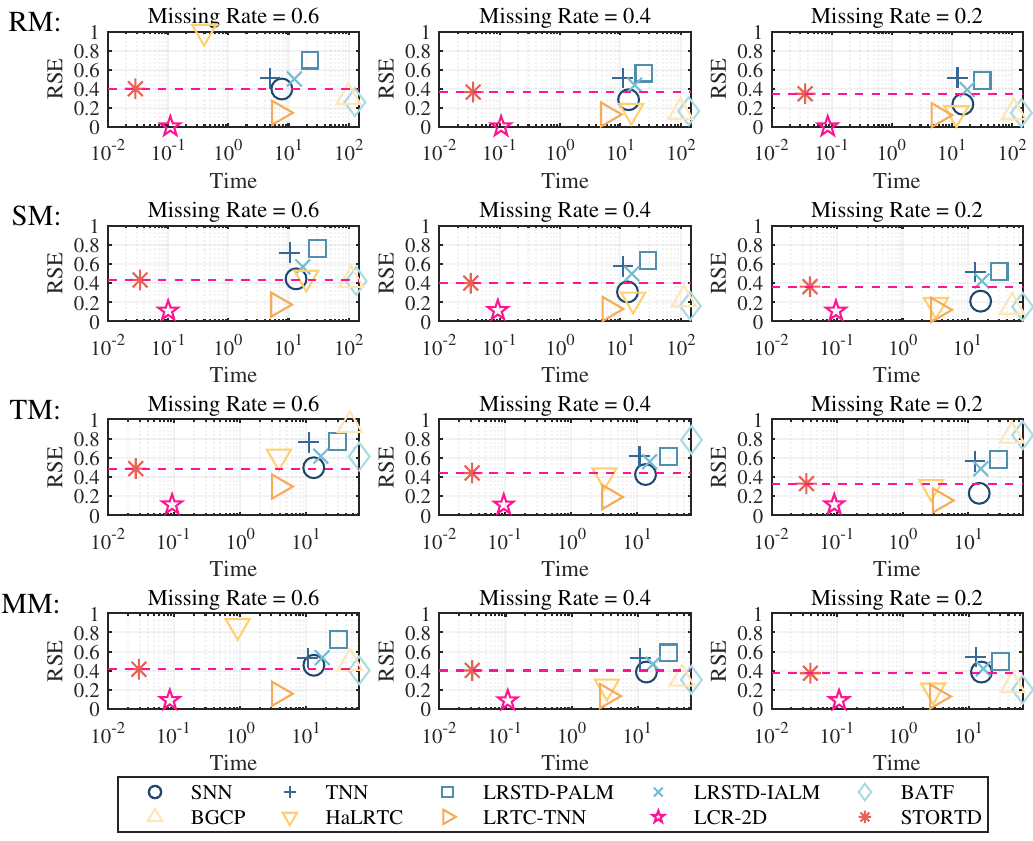}
  \caption{Comparison with offline algorithms on the Hangzhou dataset}
  \label{hz_batch}
\end{figure*}

The recovery results on the Hangzhou dataset are presented in Figure~\ref{hz_batch}, with additional results for the Guangzhou and PeMS datasets provided in Fig.3 and Fig.4 in the supplementary material. Based on these results, we summarize the following key observations:
\begin{enumerate}
\item Despite the inherent trade-off between efficiency and accuracy in online methods, our proposed approach remains highly competitive with most batch-based baselines, thanks to the incorporation of both global and local spatio-temporal structures.

\item Compared to most batch-based algorithms, our method achieves up to three orders of magnitude faster processing speed. This significant improvement in efficiency highlights its practical suitability for real-world ITS that demand timely data processing and rapid response capabilities.
\end{enumerate}

\subsection{Imputaion Result Analysis}
\begin{figure}[htbp]
  \centering
  \captionsetup[subfigure]{font=footnotesize}
  
  \begin{subfigure}[b]{1\columnwidth}
    \centering
    \includegraphics[width=\linewidth]{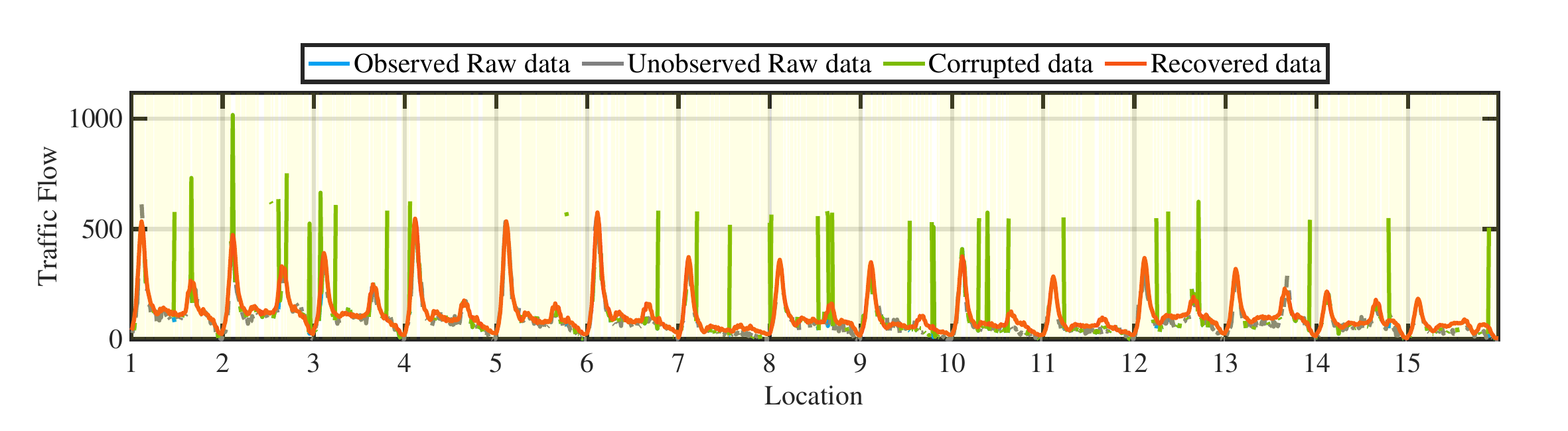}
    \caption{RM}\label{hzvs:sub1}
  \end{subfigure}
  
  \vspace{0.3cm}
  
  \begin{subfigure}[b]{1\columnwidth}
    \centering
    \includegraphics[width=\linewidth]{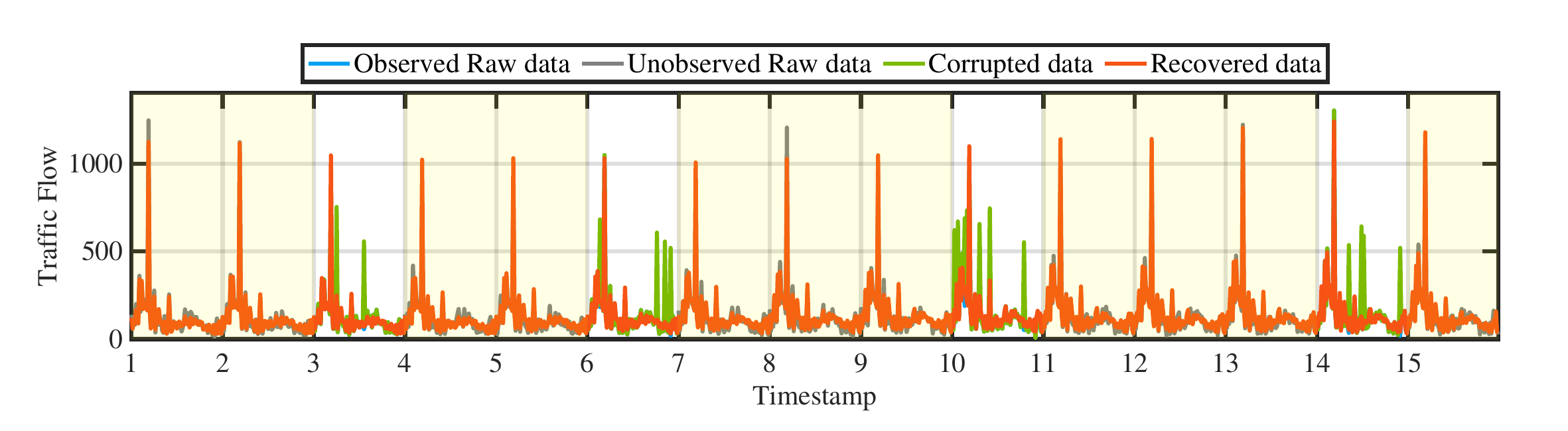}
    \caption{TM}\label{hzvs:sub2}
  \end{subfigure}
  
  \vspace{0.3cm}
  
  \begin{subfigure}[b]{1\columnwidth}
    \centering
    \includegraphics[width=\linewidth]{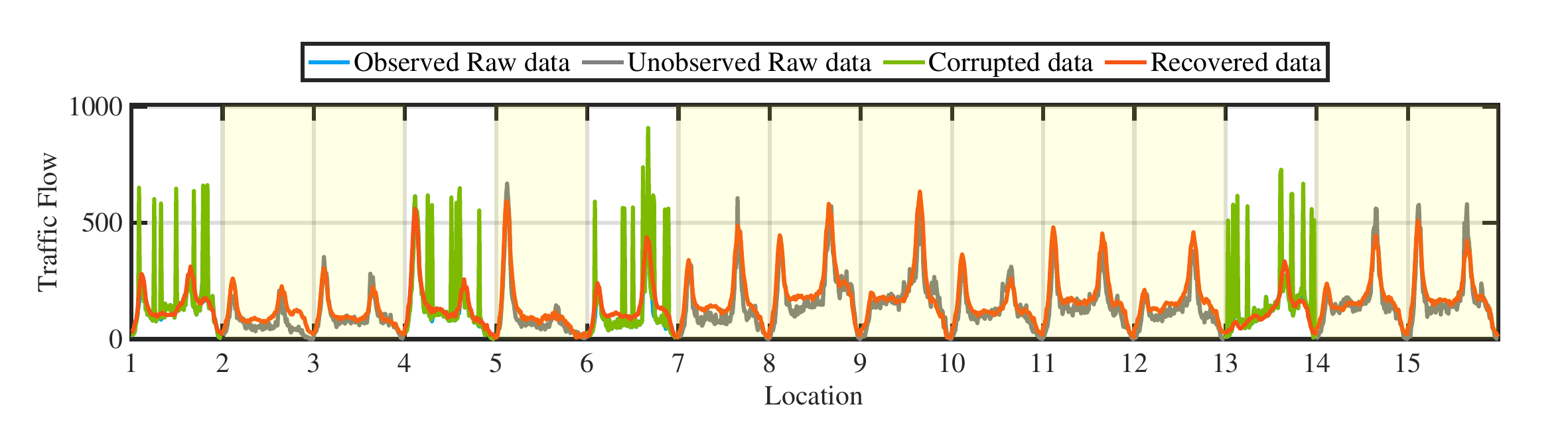}
    \caption{SM}\label{hzvs:sub3}
  \end{subfigure}
  
  \vspace{0.3cm}
  
  \begin{subfigure}[b]{1\columnwidth}
    \centering
    \includegraphics[width=\linewidth]{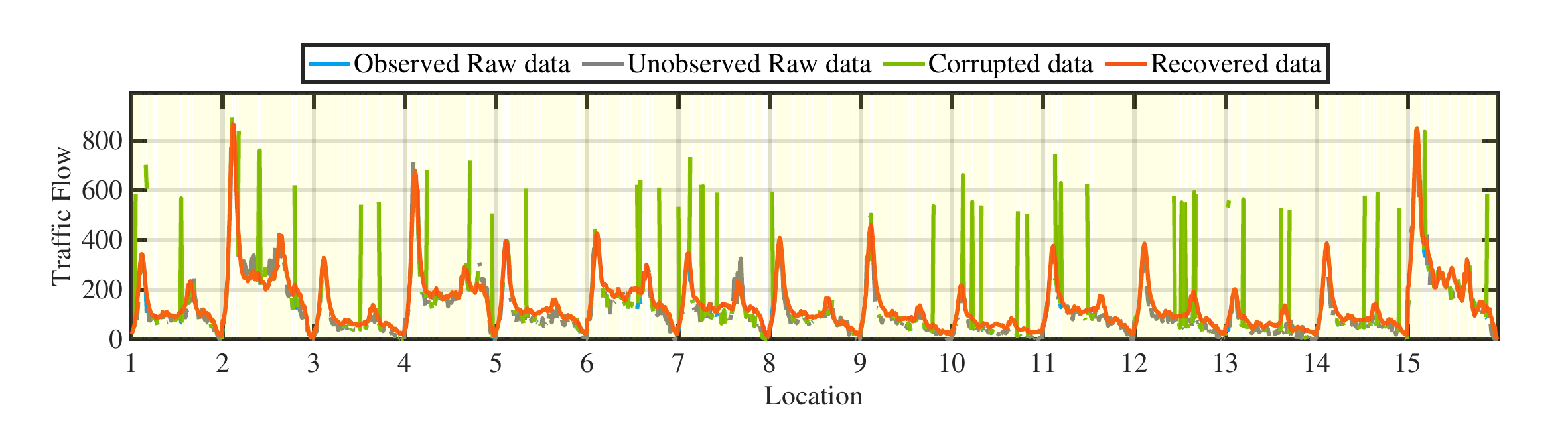}
    \caption{MM}\label{hzvs:sub4}
  \end{subfigure}
  
  \caption{Visualisation of the recovery data (represented by the red curve) and the actual data (shown as the grey curve) using the STORTD model for Hangzhou datasets with 60\% of the data missing.}
  \label{hzvs}
\end{figure}

Figure \ref{hzvs} presents representative imputation results obtained by STORTD on the Hangzhou dataset under various missing data scenarios. In the figure, the green line represents the corrupted input containing both missing values and outliers, the yellow region highlights the portions of data that are missing, and the grey line denotes the ground truth of the missing entries. Detailed imputation visualizations for the other two datasets are provided in Fig.5 and Fig.6 in the supplementary material. For clarity, we selected the traffic data from the final day of each dataset. To illustrate the effects of different missing patterns, we visualized the entire time series recorded by sensors at 15 distinct locations under RM, SM, and MM scenarios. For TM, we instead selected all sensor data within 15 consecutive timestamps to reflect the temporal aspect of structured missingness. From Figure \ref{hzvs}, it is evident that our method consistently achieves accurate recovery across diverse missing patterns, despite the presence of outliers and high missing rates, further validating its robustness and reliability in complex traffic environments.

\section{Conclusion}
In this paper, we propose STORTD, a novel online robust tensor recovery framework for spatio-temporal traffic data imputation in large-scale, dynamic ITS environments. Unlike traditional batch-based methods that repeatedly process the full dataset, STORTD operates in an online, incremental manner, significantly enhancing computational efficiency while maintaining high recovery accuracy. To fully exploit the spatio-temporal structure of the streaming traffic data, STORTD combines Tucker decomposition for global low-rank modeling with spatial and temporal regularization that captures local consistency. Such a comprehensive modeling strategy significantly enhances the robustness and adaptability of the model, demonstrating clear advantages in handling complex and diverse real-world ITS scenarios, including high missing rates, structured missing patterns, and the presence of anomalies.  Extensive experiments on three real-world traffic datasets demonstrate that STORTD outperforms state-of-the-art online and batch imputation methods in terms of accuracy, efficiency, and robustness, particularly under challenging conditions. Notably, it achieves comparable or even better recovery performance than batch-based algorithms while being up to three orders of magnitude faster, making it highly suitable for real-time ITS applications.

Although the proposed STORTD model demonstrates promising results, several directions remain open for future exploration.
First, the tensor rank in the Tucker decomposition is manually specified, which may limit adaptability across datasets with varying characteristics. Future work could consider automatic rank determination methods, such as the variational Bayesian approach \cite{zhang2018variational}, to better capture intrinsic low-rank structures.
Second, inspired by ADMM-unfolding frameworks \cite{yang2018admm}, integrating model-driven priors with deep neural networks may provide a powerful hybrid approach to further enhance recovery accuracy while preserving computational efficiency.

\bibliographystyle{IEEEtran}  %
\bibliography{ref}            

\begin{IEEEbiography}[{\includegraphics[width=0.75in,height=1.0in,clip]{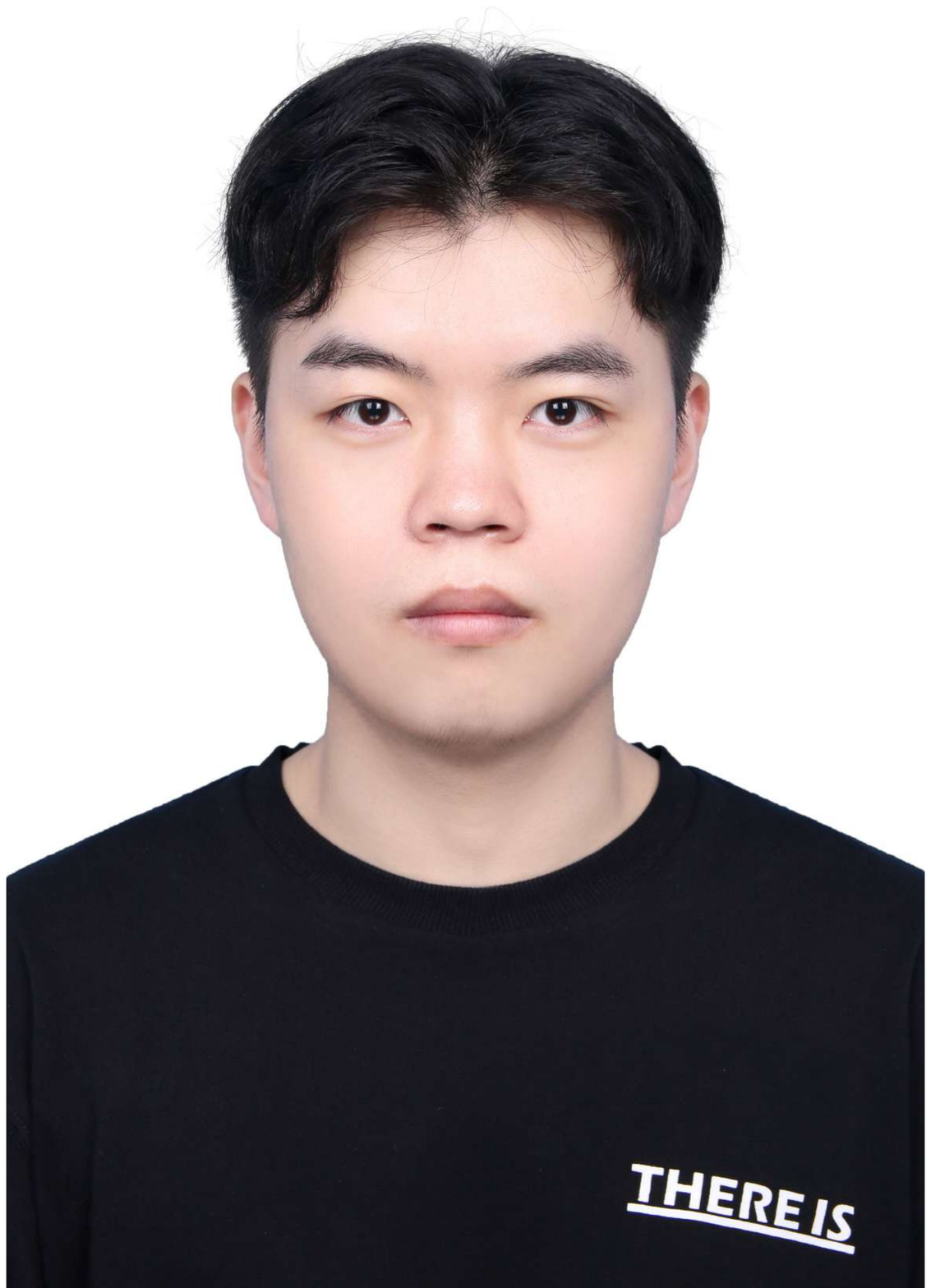}}]{Yiyang Yang} is currently a Ph.D. student at the Center for Intelligent Decision-Making and Machine Learning, School of Management, Xi'an Jiaotong University. 
His current research interests focus on high-dimensional data analysis.\end{IEEEbiography}

\begin{IEEEbiography}[{\includegraphics[width=0.75in,height=1.0in,clip]{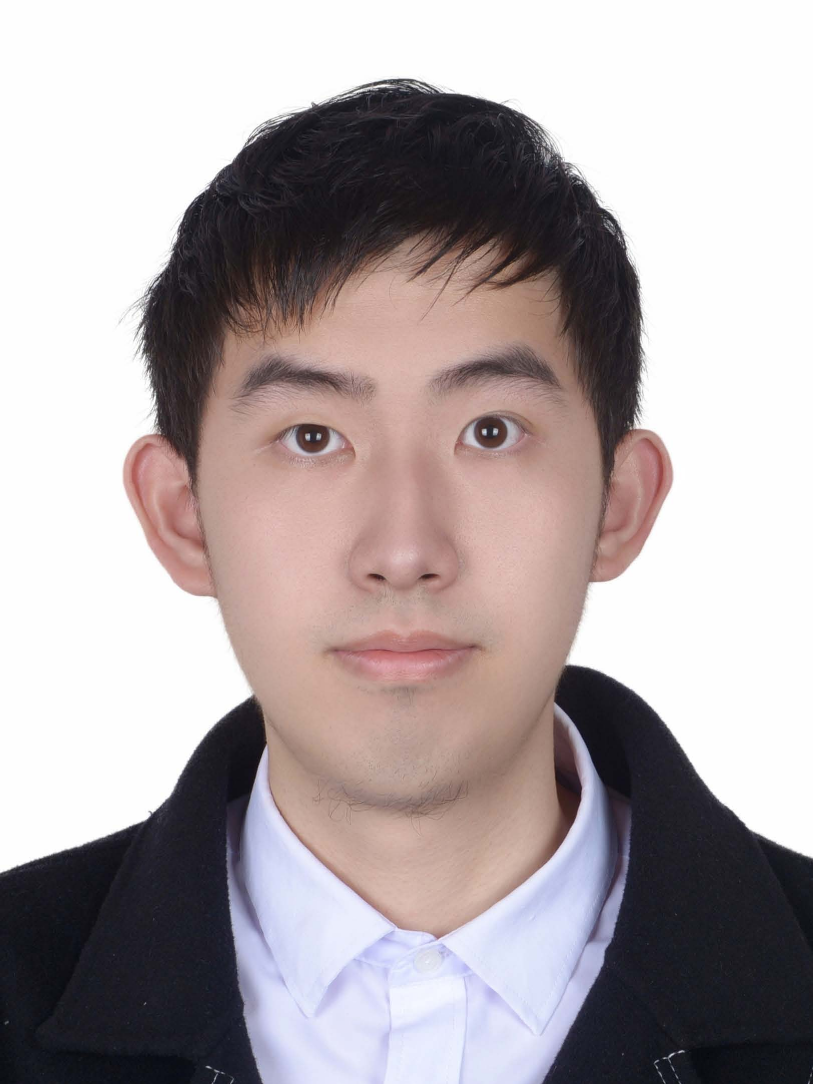}}]{Xiejian Chi} is currently an M.S. student at the Center for Intelligent Decision-Making and Machine Learning, School of Management, Xi'an Jiaotong University. 
His current research interests focus on data-driven decision making.\end{IEEEbiography}

\begin{IEEEbiography}
[{\includegraphics[width=0.75in,height=1.0in,clip]{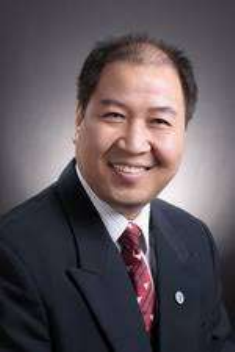}}]{Shanxing Gao} received his Ph.D. from Xi’an Jiaotong University. 

He is currently a professor in the School of Management at Xi’an Jiaotong University. His research interests include entrepreneurship, technological innovation, and intellectual property management. \end{IEEEbiography}

\begin{IEEEbiography}[{\includegraphics[width=0.75in,height=1.0in,clip, keepaspectratio]{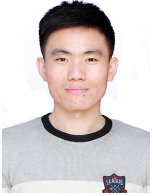}}]{Kaidong Wang} received the Ph.D. degree in applied mathematics from Xi’an Jiaotong University, Xi’an, China, in 2020.

He is currently an associate professor with the School of Management, Xi’an Jiaotong University. His current research interests include machine learning, high-dimensional data analysis, and data-driven intelligent decision-making.\end{IEEEbiography}

\begin{IEEEbiography}[{\includegraphics[width=0.75in,height=1.0in,clip, keepaspectratio]{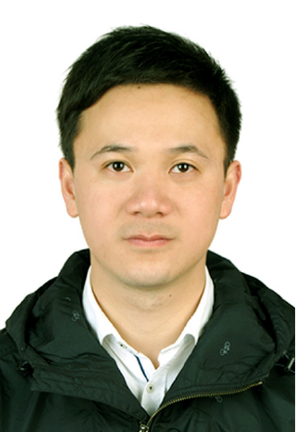}}]{Yao Wang} received the Ph.D. degree in applied mathematics from Xi’an Jiaotong University, Xi’an, China, in 2014.

He is currently a professor with the School of Management, Xi’an Jiaotong University. His current research interests include statistical signal processing, high-dimensional data analysis, and machine learning.\end{IEEEbiography}

\end{document}


\title{Supplementary Material for ``A Spatio-Temporal Online Robust Tensor Recovery Approach for Streaming Traffic Data Imputation"}

\author{Yiyang Yang, Xiejian Chi, Shanxing Gao, Kaidong Wang\textsuperscript{*}, and Yao Wang\textsuperscript{*}
}

\maketitle

\begin{abstract}
\noindent 
This supplementary material provides additional three visualization-centric results to support the main paper: (1) \textbf{online} comparative plots on \textit{Guangzhou} and \textit{PeMS}, (2) \textbf{offline (batch)} comparative plots on \textit{Guangzhou} and \textit{PeMS}, and (3) \textbf{qualitative imputation} visualizations under four missing patterns: Random Missing (RM), Temporal Missing (TM), Spatial Missing (SM) and Mix Missing (MM). Together, these visuals substantiate the robustness and accuracy trends reported in the main text.
\end{abstract}
\section{Online Algorithms on Guangzhou and PeMS}
\label{sec:online-gz-pems}

\begin{figure}[htbp]
  \centering
  \captionsetup[subfigure]{font=footnotesize}
  \begin{subfigure}[b]{0.48\columnwidth}
    \centering\includegraphics[width=\linewidth]{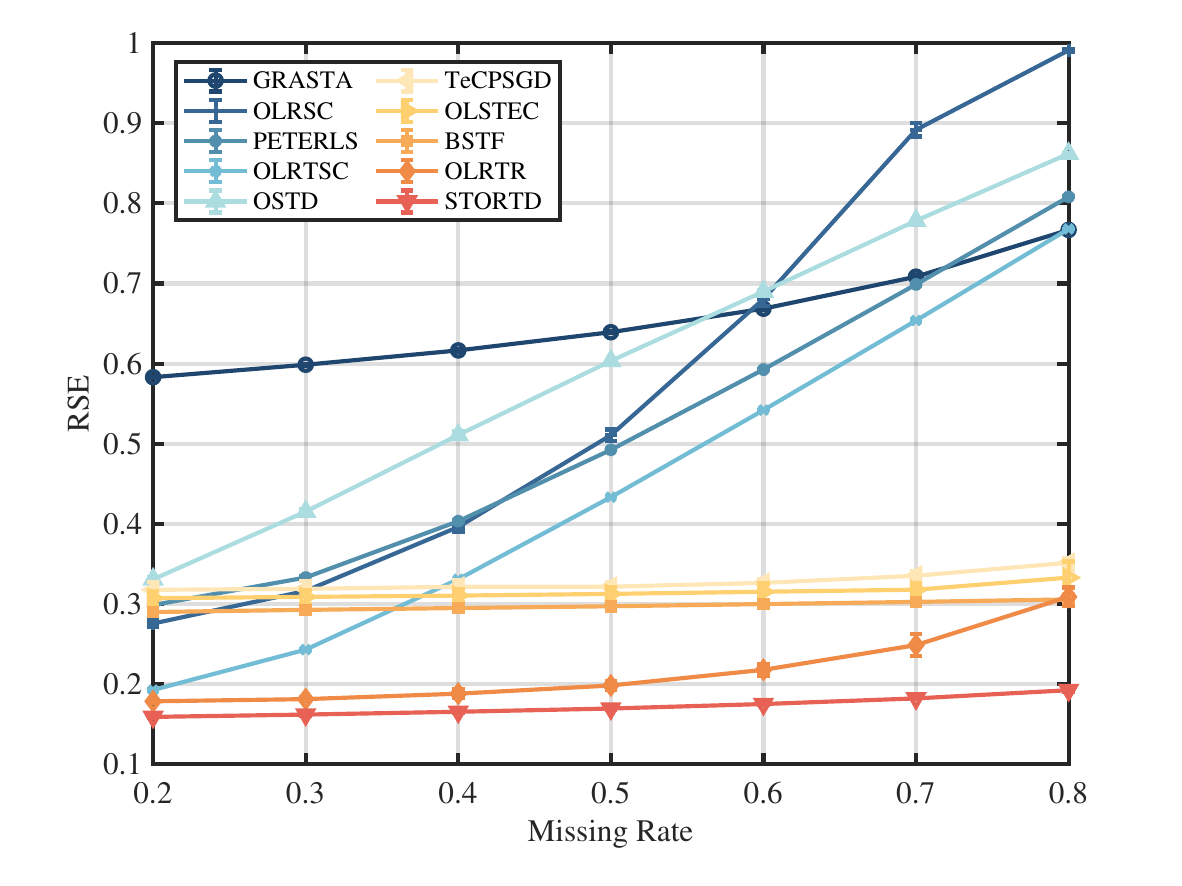}
    \caption{RM}
  \end{subfigure}\hfill
  \begin{subfigure}[b]{0.48\columnwidth}
    \centering\includegraphics[width=\linewidth]{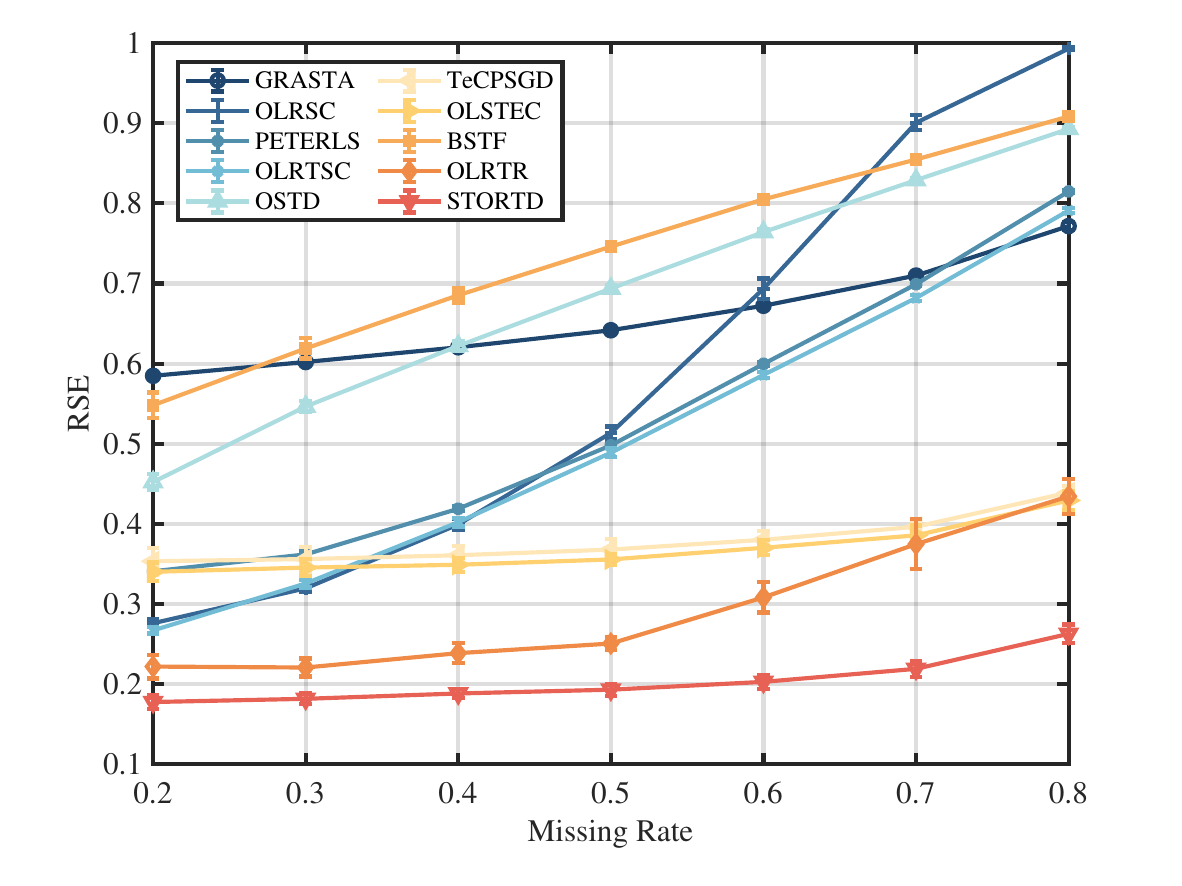}
    \caption{TM}
  \end{subfigure}

  \medskip

  \begin{subfigure}[b]{0.48\columnwidth}
    \centering\includegraphics[width=\linewidth]{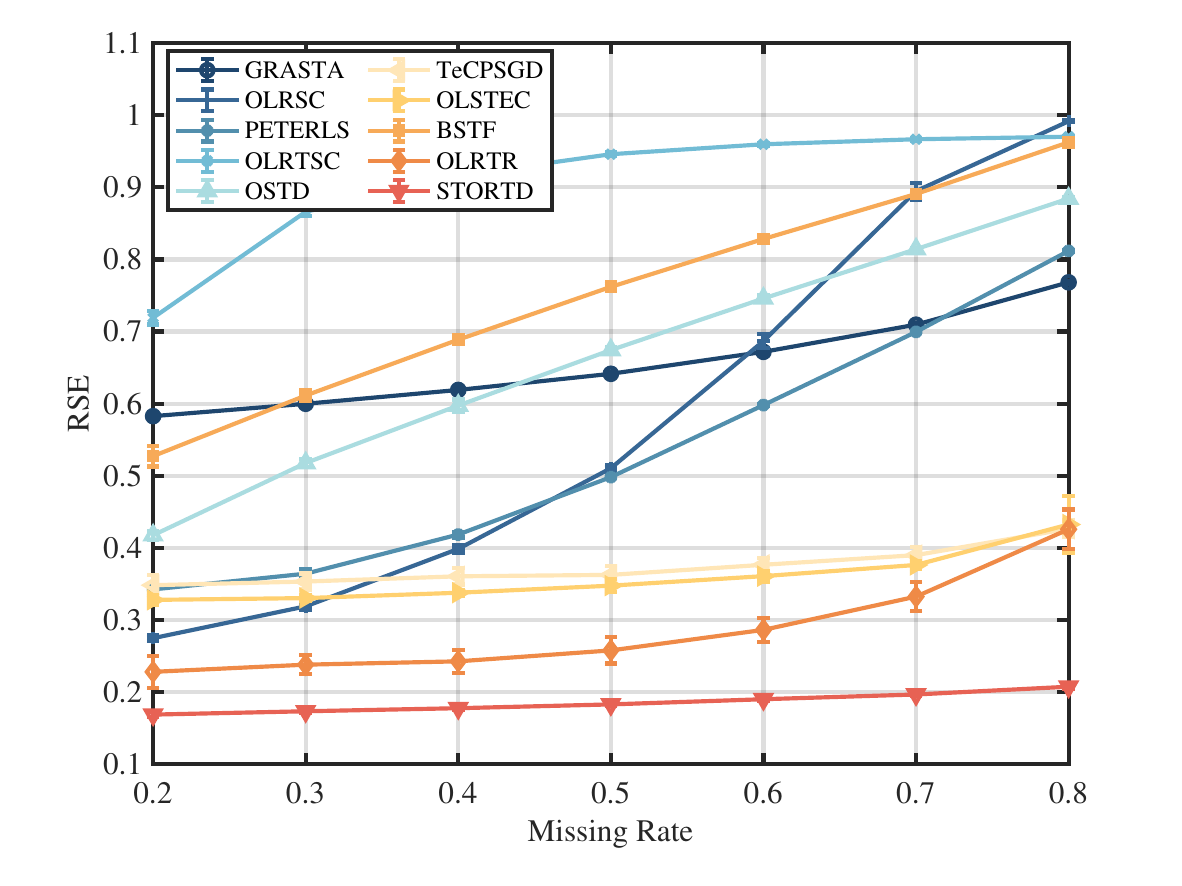}
    \caption{SM}
  \end{subfigure}\hfill
  \begin{subfigure}[b]{0.48\columnwidth}
    \centering\includegraphics[width=\linewidth]{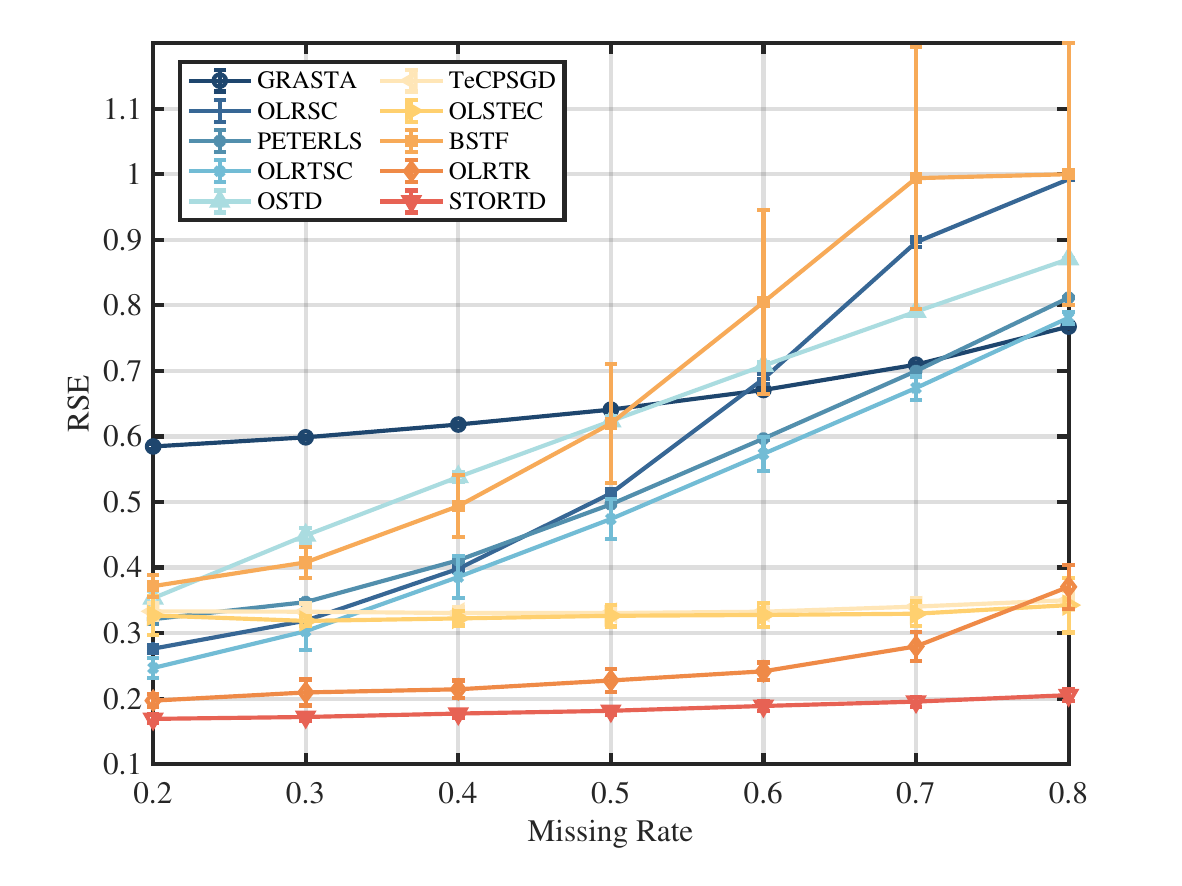}
    \caption{MM}
  \end{subfigure}
  \caption{Comparison with online algorithms on the \textit{Guangzhou} dataset.}
  \label{fig:gz_online}
\end{figure}

Fig.~\ref{fig:gz_online} reports the comparison results of online algorithms on the \textit{Guangzhou} dataset. Across RM/TM/SM/MM and missing rates from $0.2$ to $0.8$, STORTD attains the lowest RSE in all configurations. The advantage becomes more pronounced under structured missingness, evidencing the benefit of the combined spatial (graph-Laplacian) and temporal (Toeplitz) regularization. At the highest missingness levels, competing online methods deteriorate substantially faster than STORTD.

\begin{figure}[t]
  \centering
  \captionsetup[subfigure]{font=footnotesize}
  \begin{subfigure}[b]{0.48\columnwidth}
    \centering\includegraphics[width=\linewidth]{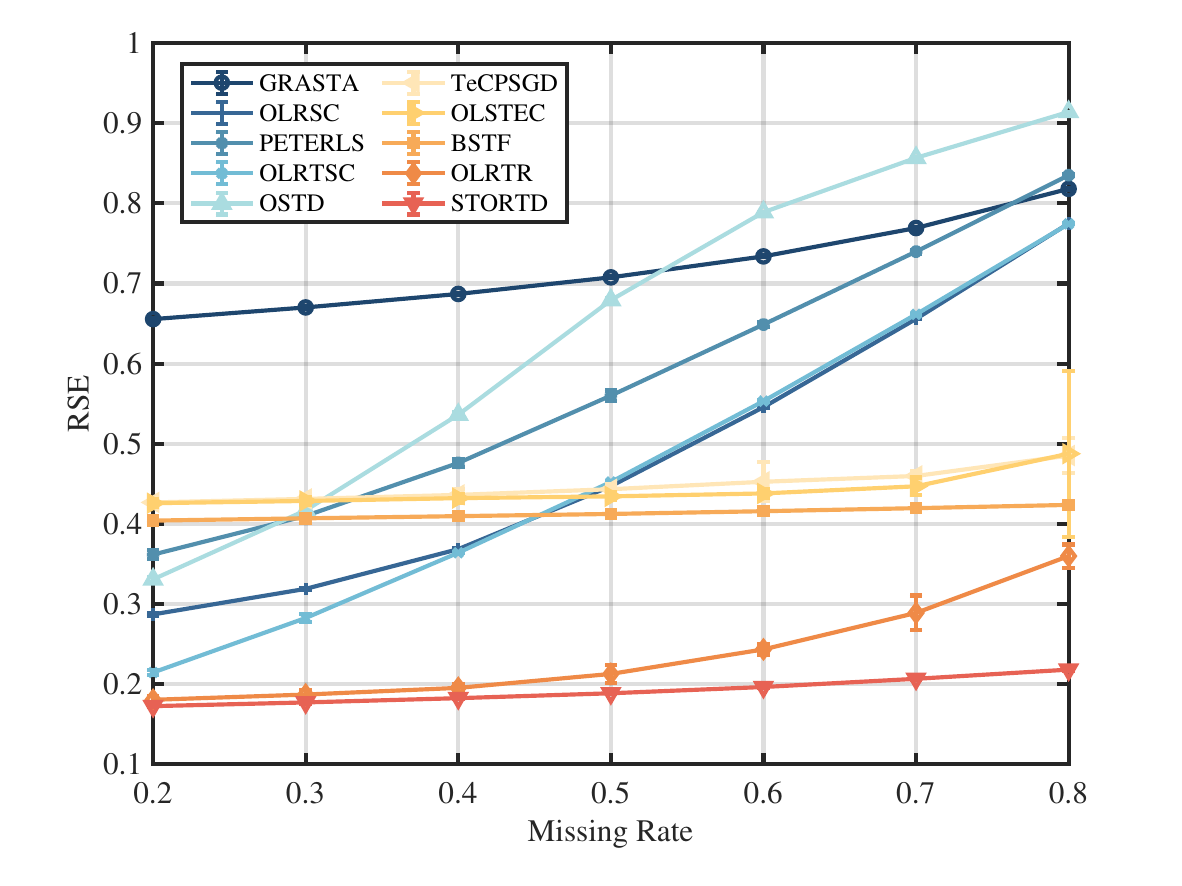}
    \caption{RM}
  \end{subfigure}\hfill
  \begin{subfigure}[b]{0.48\columnwidth}
    \centering\includegraphics[width=\linewidth]{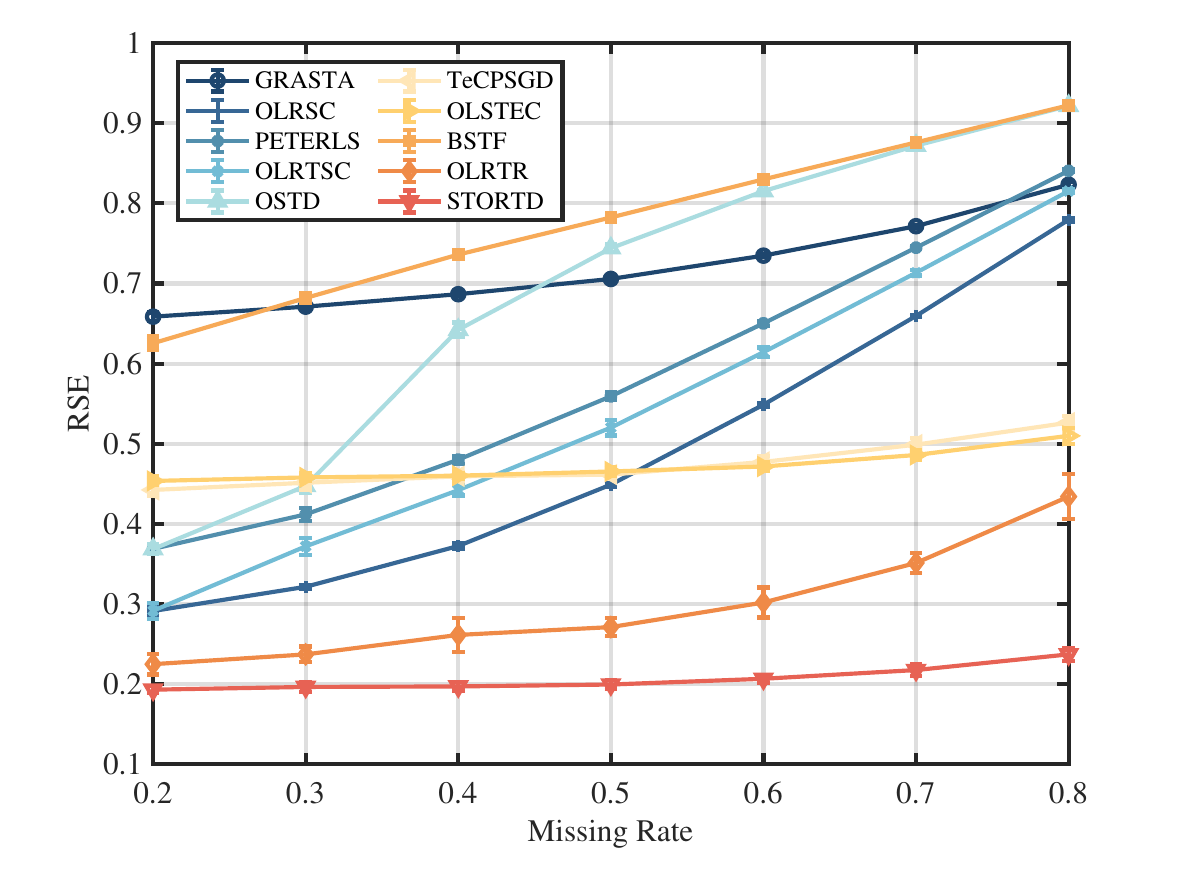}
    \caption{TM}
  \end{subfigure}

  \medskip

  \begin{subfigure}[b]{0.48\columnwidth}
    \centering\includegraphics[width=\linewidth]{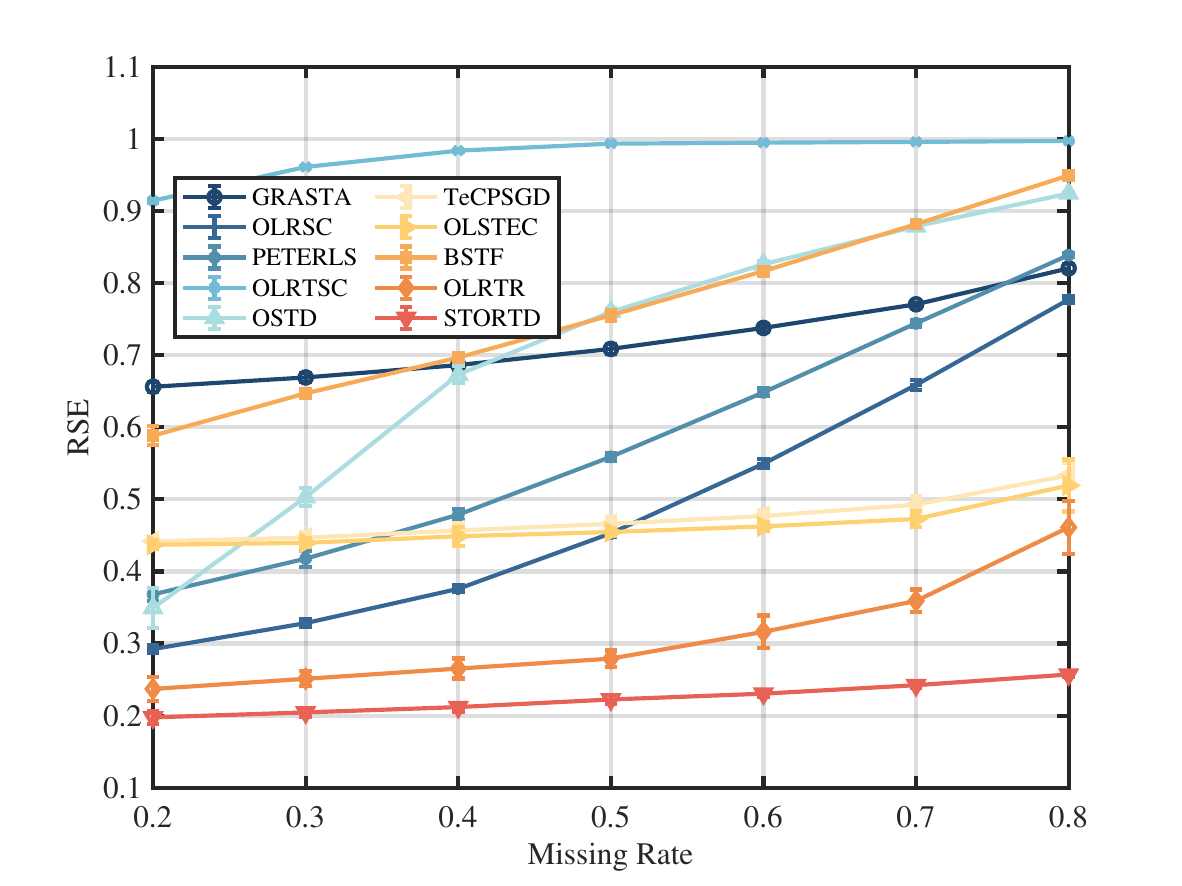}
    \caption{SM}
  \end{subfigure}\hfill
  \begin{subfigure}[b]{0.48\columnwidth}
    \centering\includegraphics[width=\linewidth]{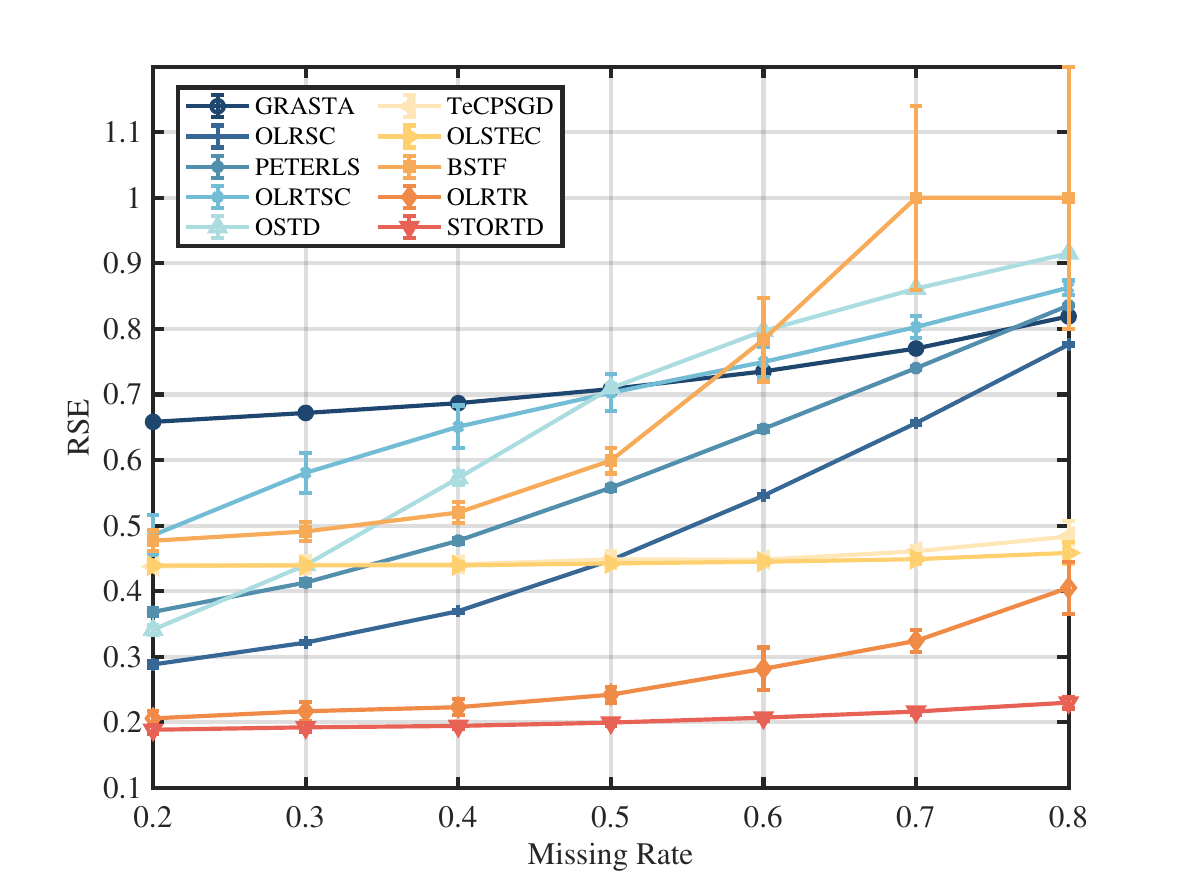}
    \caption{MM}
  \end{subfigure}
  \caption{Comparison with online algorithms on the \textit{PeMS} dataset.}
  \label{fig:pems_online}
\end{figure}

With the same setting, Fig.~\ref{fig:pems_online} presents the comparison results of online algorithms on the \textit{PeMS} dataset. STORTD leads across the entire range of missingness and maintains a stable margin for four missing patterns, indicating effective exploitation of spatial adjacency and temporal continuity in streaming. As the missing rate increases, the RSE curves for STORTD grow more slowly than those of competing methods, demonstrating graceful degradation under severe data loss.

\begin{figure}[t]
  \centering
  \includegraphics[width=\columnwidth]{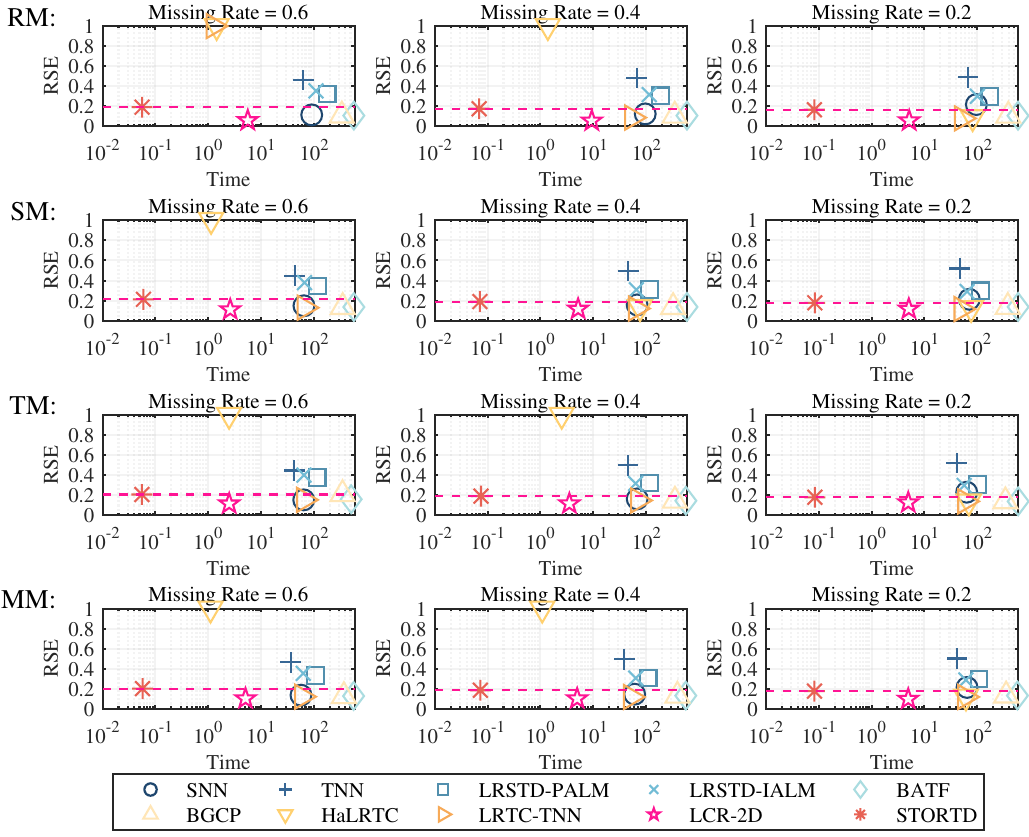}
  \caption{Comparison with offline algorithms on the \textit{Guangzhou} dataset.}
  \label{fig:gz_batch}
\end{figure}

\section{Offline (Batch) Algorithms on Guangzhou and PeMS}\label{sec:offline-gz-pems}

Fig.~\ref{fig:gz_batch} summarizes the comparison results of offline algorithms on the \textit{Guangzhou} dataset. Across RM/TM/SM/MM and three missing rates (0.2, 0.4, 0.6), although offline methods benefit from full-tensor access and longer runtimes, STORTD remains competitive across missingness levels and attains favorable RSE--time trade-offs.\\

\begin{figure}[H]
  \centering
  \includegraphics[width=\columnwidth]{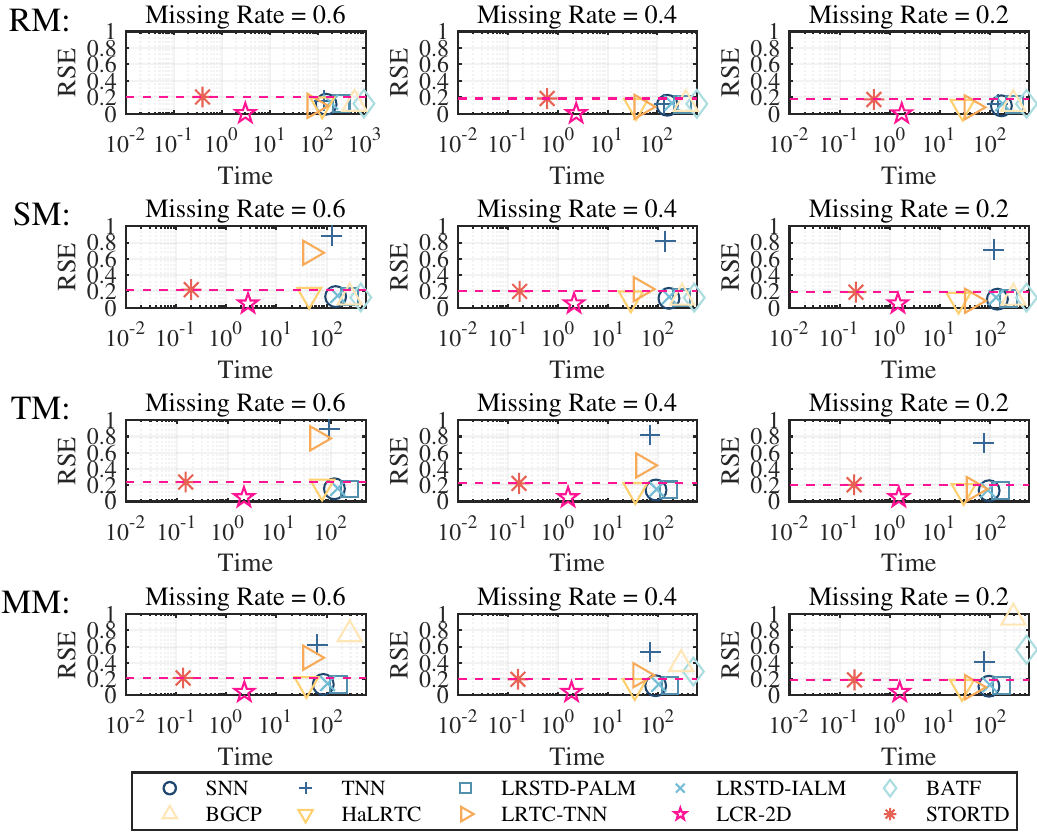}
  \caption{Comparison with offline algorithms on the \textit{PeMS} dataset.}
  \label{fig:pems_batch}
\end{figure}

With the same setting, Fig.~\ref{fig:pems_batch} shows a similar comparison results of offline algorithms holds on the \textit{PeMS} dataset, that STORTD traces an efficient frontier balancing accuracy and latency.

\begin{figure}[htpb]
  \centering
  \captionsetup[subfigure]{font=footnotesize}
  \begin{subfigure}[b]{0.98\columnwidth}
    \centering\includegraphics[width=\linewidth]{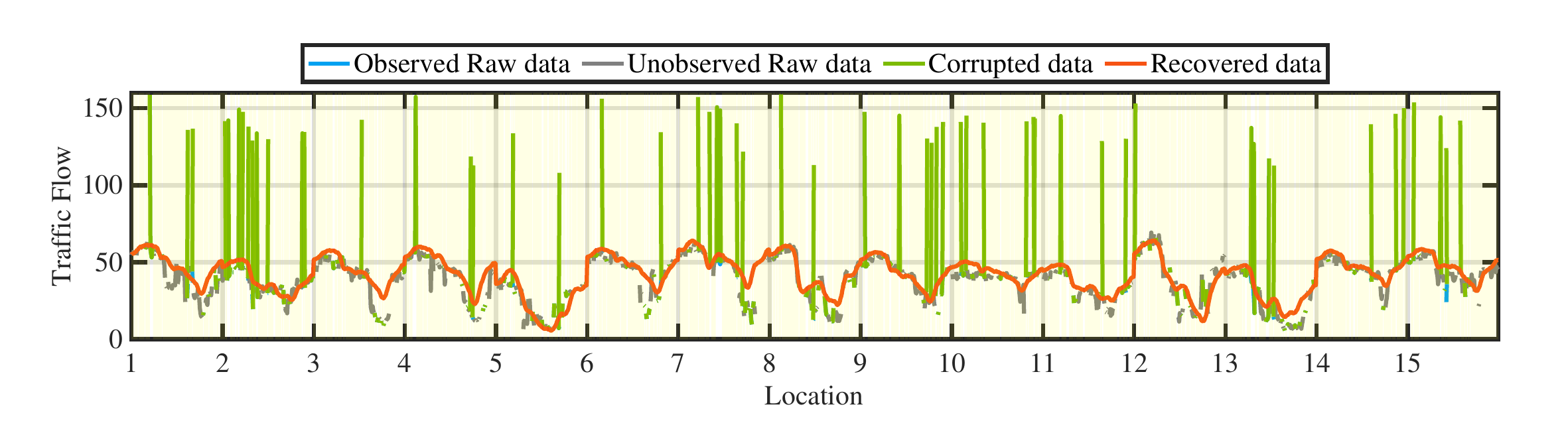}
    \caption{RM}
  \end{subfigure}

  \vspace{0.15cm}

  \begin{subfigure}[b]{0.98\columnwidth}
    \centering\includegraphics[width=\linewidth]{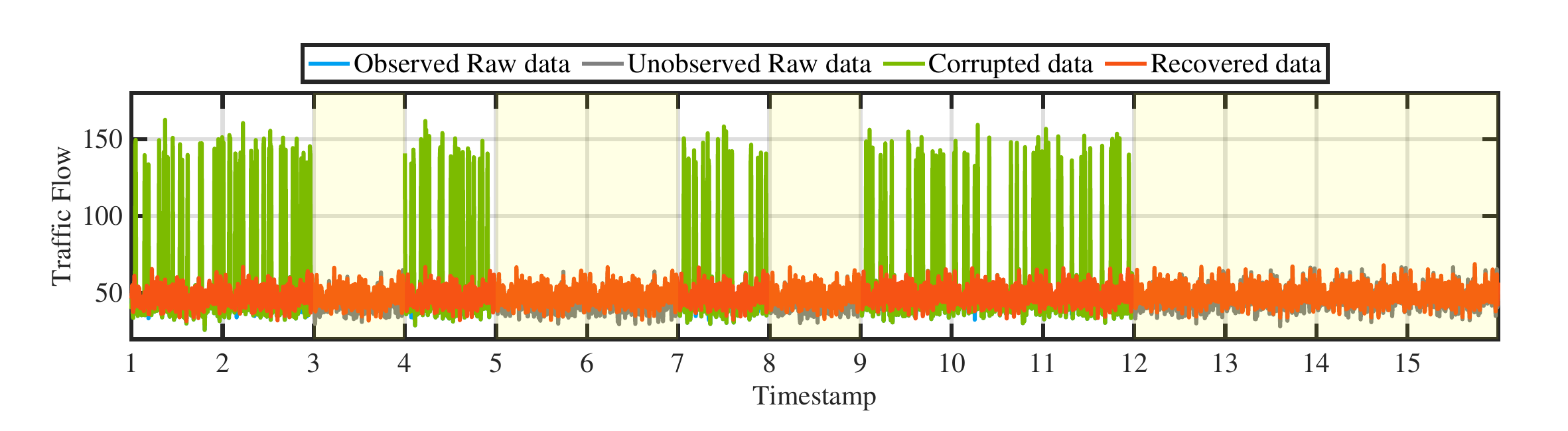}
    \caption{TM}
  \end{subfigure}

  \vspace{0.15cm}

  \begin{subfigure}[b]{0.98\columnwidth}
    \centering\includegraphics[width=\linewidth]{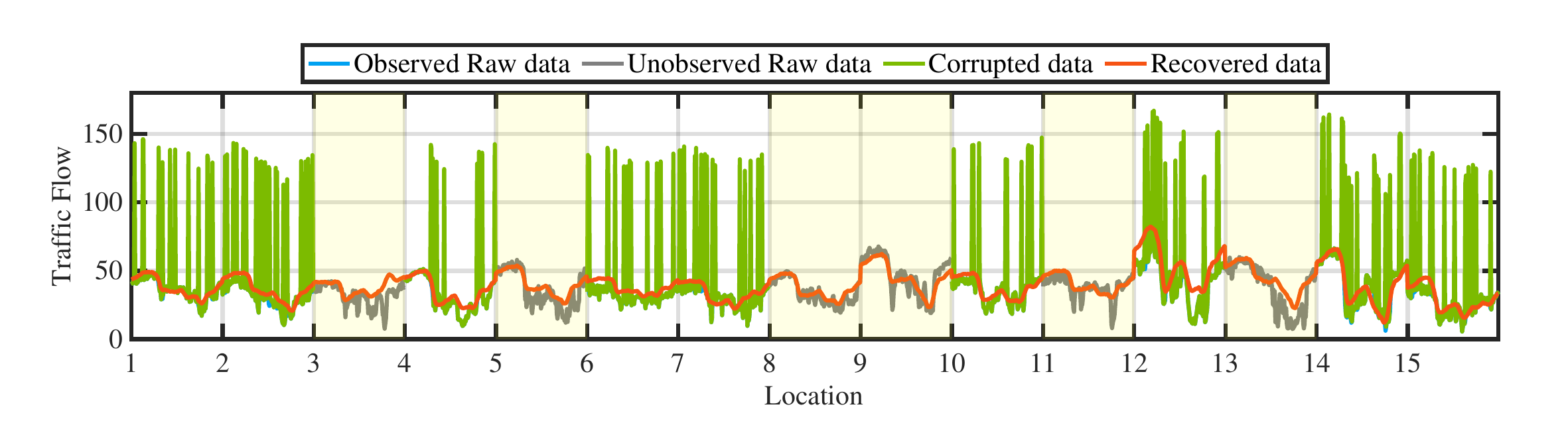}
    \caption{SM}
  \end{subfigure}

  \vspace{0.15cm}

  \begin{subfigure}[b]{0.98\columnwidth}
    \centering\includegraphics[width=\linewidth]{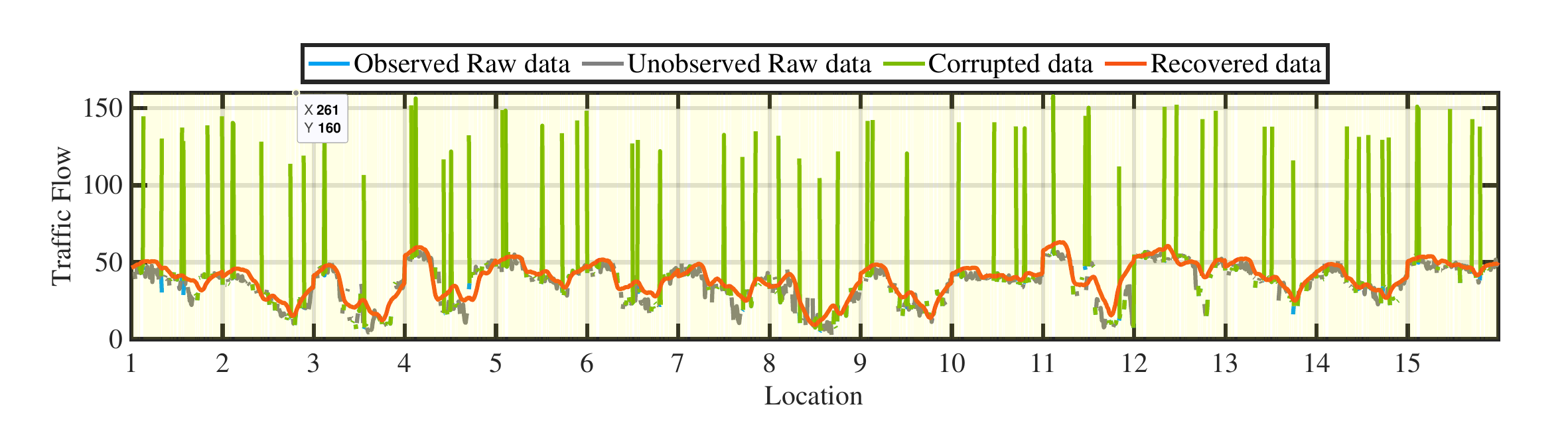}
    \caption{MM}
  \end{subfigure}

  \caption{STORTD recovery (red) vs.\ ground truth (grey) on the \textit{Guangzhou} dataset with $60\%$ missingness.}
  \label{fig:gz_viz}
\end{figure}

\section{QUALITATIVE IMPUTATION VISUALIZATIONS on Guangzhou and PeMS}\label{sec:viz-gz-pems}
\begin{figure}[H]
  \centering
  \captionsetup[subfigure]{font=footnotesize}
  \begin{subfigure}[b]{0.98\columnwidth}
    \centering\includegraphics[width=\linewidth]{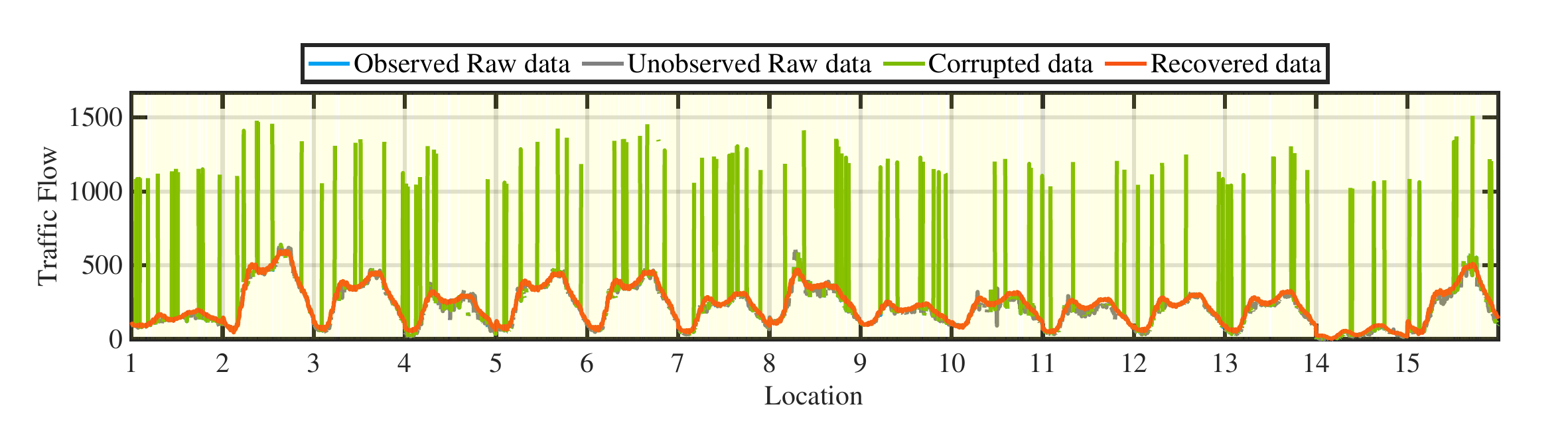}
    \caption{RM}
  \end{subfigure}

  \vspace{0.15cm}

  \begin{subfigure}[b]{0.98\columnwidth}
    \centering\includegraphics[width=\linewidth]{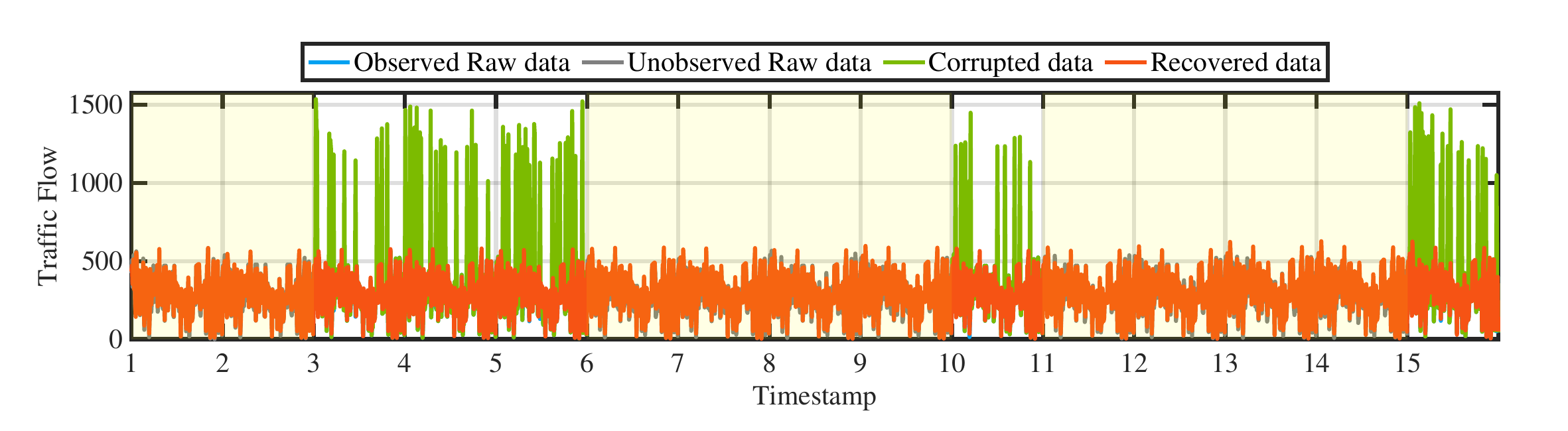}
    \caption{TM}
  \end{subfigure}

  \vspace{0.15cm}

  \begin{subfigure}[b]{0.98\columnwidth}
    \centering\includegraphics[width=\linewidth]{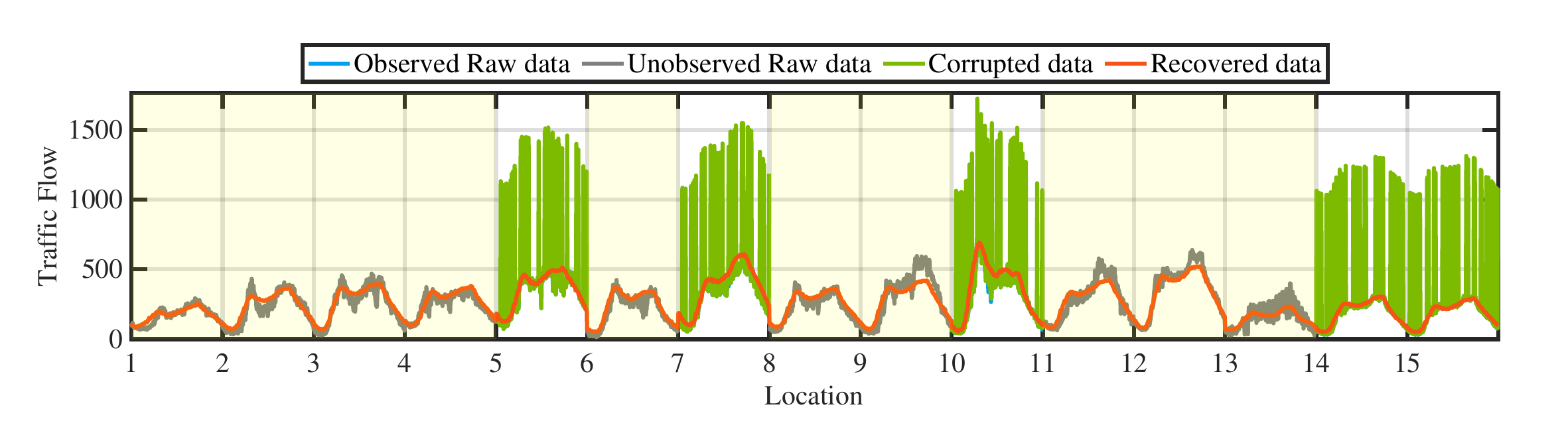}
    \caption{SM}
  \end{subfigure}

  \vspace{0.15cm}

  \begin{subfigure}[b]{0.98\columnwidth}
    \centering\includegraphics[width=\linewidth]{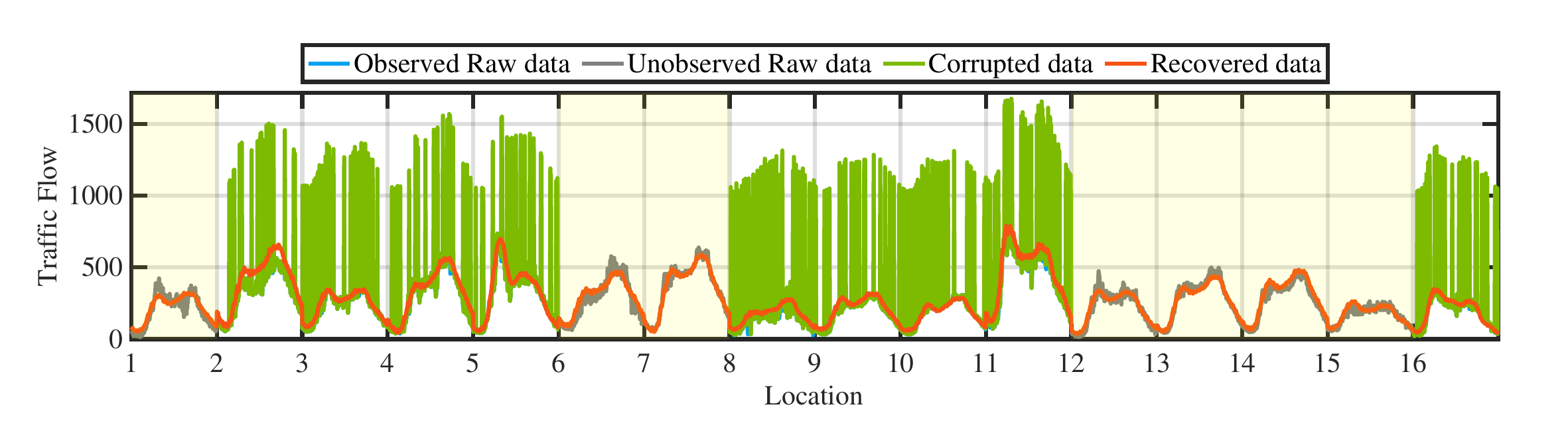}
    \caption{MM}
  \end{subfigure}

  \caption{STORTD recovery (red) vs.\ ground truth (grey) on the \textit{PeMS} dataset with $60\%$ missingness.}
  
  \label{fig:pems_viz}
\end{figure}

We visualize STORTD’s recovered series against ground truth at $60\%$ missingness across RM/TM/SM/MM (see Fig.~\ref{fig:gz_viz} and Fig.~\ref{fig:pems_viz}). On the both \textit{Guangzhou} and \textit{PeMS} datasets, the recovered curves closely follow the true curves. Although the four missing patterns introduce localized and structured gaps, STORTD reconstructs these segments coherently by fusing spatial neighbors (via graph Laplacian coupling) and adjacent temporal correlation (via Toeplitz differences), while suppressing outliers.

\bibliographystyle{IEEEtran}